\documentclass[10pt,twocolumn,letterpaper]{article}
\usepackage[pagenumbers]{include/cvpr/cvpr} 
\usepackage{graphicx}
\usepackage{amsmath}
\usepackage{amssymb}
\usepackage{booktabs}
\usepackage{inconsolata}
\usepackage{multirow}
\usepackage{enumitem}
\usepackage{algorithm}
\usepackage{listings}
\usepackage{float}
\usepackage{xcolor, colortbl}
\usepackage{hhline}
\usepackage[pagebackref,breaklinks,colorlinks]{hyperref}
\usepackage[capitalize]{cleveref} 
\crefname{section}{Sec.}{Secs.}
\Crefname{section}{Section}{Sections}
\Crefname{table}{Table}{Tables}
\crefname{table}{Tab.}{Tabs.}

\def\cvprPaperID{***} 
\def\confName{CVPR}
\def\confYear{2023}
\newcommand{\tbf}[1]{\textbf{#1}}
\newcommand{\tsc}[1]{\textsuperscript{#1}}
\newcommand{\mulccol}[1]{\multicolumn{2}{c}{#1}}

\title{Hierarchical discriminative learning improves visual representations of biomedical microscopy}

\author{%
  Cheng Jiang\tsc{1*}\quad
  Xinhai Hou\tsc{1*}\quad
  Akhil Kondepudi\tsc{1}\quad
  Asadur Chowdury\tsc{1}\quad\\
  Christian W. Freudiger\tsc{2}\quad
  Daniel A. Orringer\tsc{3}\quad
  Honglak Lee\tsc{1}\quad
  Todd C. Hollon\tsc{1}\\[1em]
  \tsc{1}University of Michigan\quad
  \tsc{2}Invenio Imaging\quad
  \tsc{3}New York University\quad
  \tsc{*}Equal Contribution\\[1em]
  \texttt{\{chengjia, xinhaih, tocho\}@umich.edu}\quad
  \url{https://hidisc.mlins.org/}
}
\begin{document}
\maketitle
\begin{abstract}
Learning high-quality, self-supervised, visual representations is essential to advance the role of computer vision in biomedical microscopy and clinical medicine. Previous work has focused on self-supervised representation learning (SSL) methods developed for instance discrimination and applied them directly to image patches, or fields-of-view, sampled from gigapixel whole-slide images (WSIs) used for cancer diagnosis. However, this strategy is limited because it (1) assumes patches from the same patient are independent, (2) neglects the patient-slide-patch hierarchy of clinical biomedical microscopy, and (3) requires strong data augmentations that can degrade downstream performance. Importantly, sampled patches from WSIs of a patient's tumor are a diverse set of image examples that capture the same underlying cancer diagnosis. This motivated HiDisc, a data-driven method that leverages the inherent patient-slide-patch hierarchy of clinical biomedical microscopy to define a \underline{hi}erarchical \underline{disc}riminative learning task that implicitly learns features of the underlying diagnosis. HiDisc uses a self-supervised contrastive learning framework in which positive patch pairs are defined based on a common ancestry in the data hierarchy, and a unified patch, slide, and patient discriminative learning objective is used for visual SSL. We benchmark HiDisc visual representations on two vision tasks using two biomedical microscopy datasets, and demonstrate that (1) HiDisc pretraining outperforms current state-of-the-art self-supervised pretraining methods for cancer diagnosis and genetic mutation prediction, and (2) HiDisc learns high-quality visual representations using natural patch diversity without strong data augmentations.
\end{abstract}


\begin{figure}[t!]
    \centering
    \includegraphics[width=\columnwidth]{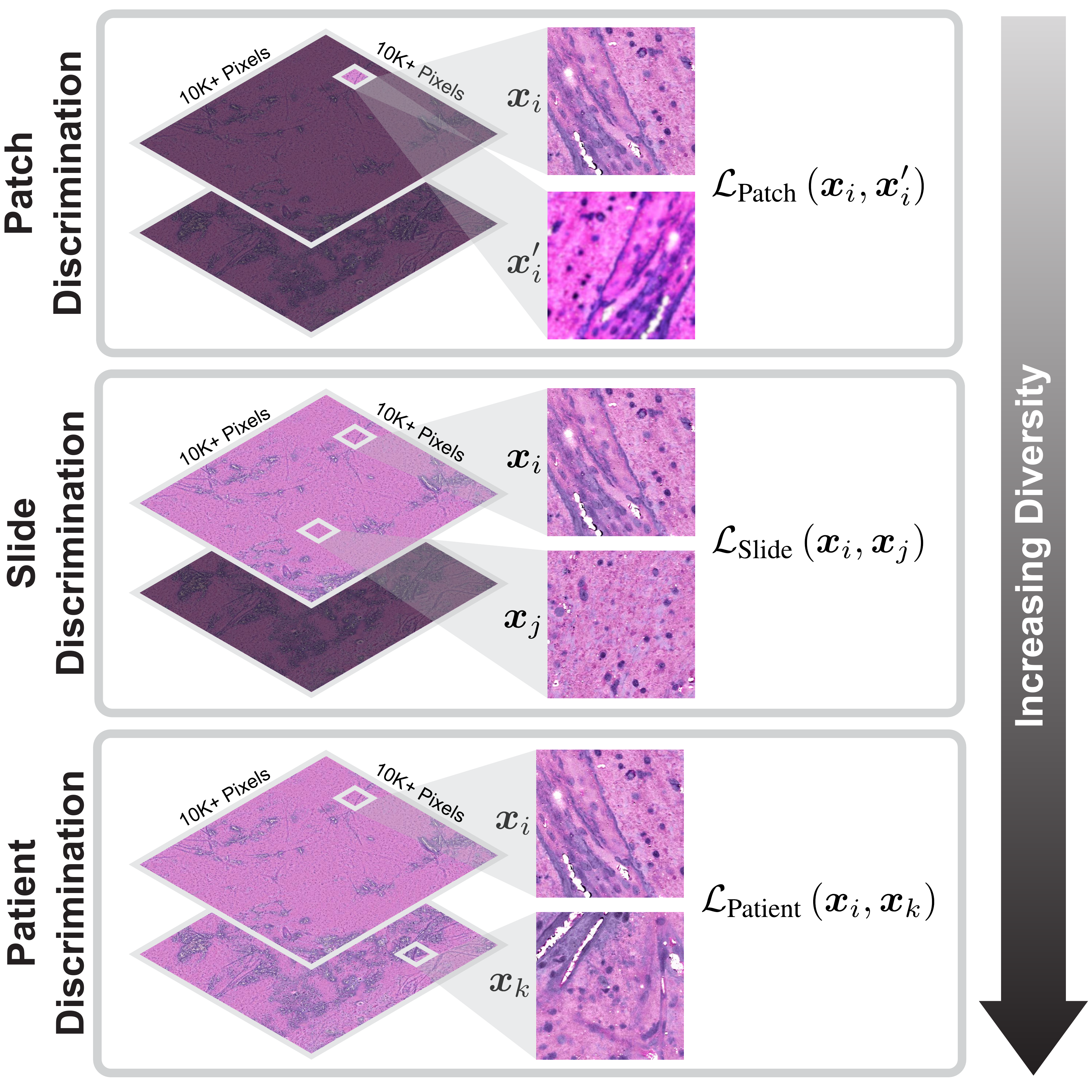}
    \caption{\textbf{Hierarchical self-supervised discriminative learning for visual representations}. Clinical biomedical microscopy has a hierarchical patch-slide-patient data structure. HiDisc combines patch, slide, and patient discrimination into a unified self-supervised learning task.}
    \label{fig:intro}
\end{figure}

\section{Introduction} \label{sec:intro}
Biomedical microscopy is an essential imaging method and diagnostic modality in biomedical research and clinical medicine. The rise of digital pathology and whole-slide images (WSIs) has increased the role of computer vision and machine learning-based approaches for analyzing microscopy data \cite{Tizhoosh2018-bo}. Improving the quality of visual representation learning of biomedical microscopy is critical to introducing decision support systems and automated diagnostic tools into clinical and laboratory medicine.

Biomedical microscopy and WSIs present several unique computer vision challenges, including that image resolutions can be large (10K$\times$10K pixels) and annotations are often limited to weak slide-level or patient-level labels. Moreover, even weak annotations are challenging to obtain in order to protect patient health information and ensure patient privacy \cite{Wiens2019-mh}. Additionally, data that predates newly developed or future clinical testing methods, such as genomic or methylation assays, also lack associated weak annotations. Because of large WSI sizes and weak annotations, the majority of computer vision research in biomedical microscopy has focused on WSI classification using a weakly supervised, patch-based, multiple instance learning (MIL) framework \cite{Lu2021-vn, Lu2021-xi, Campanella2019-oc, Hou2016-fq, Chen2022-bp, Shao2021-sj}. Patches are arbitrarily defined fields-of-view (e.g., 256$\times$256 pixels) that can be used for model  input. The classification tasks include identifying the presence of cancerous tissue, such as breast cancer metastases in lymph node biopsies \cite{Ehteshami_Bejnordi2017-fb}, differentiating specific cancer types \cite{Coudray2018-mp, Chen2022-bp, Hollon2020-ez}, predicting genetic mutations \cite{Liu2022-gc, Kather2020-fg, Coudray2018-mp}, and patient prognostication \cite{Lipkova2022-rv, Chen2022-se}. A limitation of end-to-end MIL frameworks for WSI classification is the reliance on weak annotations to train a patch feature extractor and achieve high-quality patch-level representation learning. This limitation, combined with the challenge of obtaining fully annotated, high-quality WSIs, necessitates better methods for self-supervised representation learning (SSL) of biomedical microscopy.

To date, research into improving the quality and efficiency of patch-level representation learning with\textit{out} annotations has been limited. Previous studies have focused on using known SSL methods, such as contrastive learning \cite{Schirris2022-sh, Lu2019-op, Stacke2022-ts}, and applying them directly to WSI patches for visual pretraining. These SSL methods are not optimal because the majority use instance (i.e., patch) discrimination as the pretext learning task \cite{Chen2020-td, Grill2020-at, Chen2020-gs, Caron2021-lk, Wu2018-ur}. Patches belonging to the same slide or patient are correlated, which can decrease the learning efficiency. Instance discrimination alone does not account for patches from a common slide or patient being different and diverse views of the same underlying pathology. Moreover, previous SSL methods neglect the inherent patient-slide-patch data hierarchy of clinical biomedical microscopy as shown in Figure \ref{fig:intro}. This hierarchical data structure is not used to improve representation learning when training via a standard SSL objective. Lastly, most SSL methods require strong data augmentations for instance discrimination tasks \cite{Chen2020-td}. However, strong and domain-agnostic augmentations can worsen representation learning in microscopy images by corrupting semantically important and discriminative features \cite{Hussain2017-jd, Stacke2022-ts}.

Here, we introduce a method that leverages the inherent patient-slide-patch hierarchy of clinical biomedical microscopy to define a self-supervised \underline{hi}erarchical \underline{disc}riminative learning task, called HiDisc. HiDisc uses a self-supervised contrastive learning framework such that positive patch pairs are defined based on a common ancestry in the data hierarchy, and a combined patch, slide, and patient discriminative learning objective is used for visual SSL. By sampling patches across the data hierarchy, we introduce increased diversity between the positive examples, allowing for better visual representation learning and bypassing the need for strong, out-of-domain data augmentations. While we examine the HiDisc learning objective in the context of contrastive learning, it can be generalized to any siamese representation learning method \cite{Chen2020-gs}. 

We benchmark HiDisc self-supervised pretraining on two computer vision tasks using two diverse biomedical microscopy datasets: (1) multiclass histopathologic cancer diagnosis using stimulated Raman scattering microscopy \cite{Orringer2017-nn} and (2) molecular genetic mutation prediction using light microscopy of hematoxylin and eosin (H\&E)-stained cancer specimens \cite{Liu2020-cl}. These tasks are selected because of their clinical importance and they represent examples of how deep learning-based computer vision methods can push the limits of what is achievable through biomedical microscopy \cite{Hollon2020-ez, Jiang2022-bk, Liu2020-wh, Kather2020-fg}. We benchmark HiDisc in comparison to several state-of-the-art SSL methods, including SimCLR \cite{Chen2020-td}, BYOL\cite{Grill2020-at}, and VICReg \cite{Bardes2022-md}. We demonstrate that HiDisc has superior performance compared to other SSL methods across both datasets and computer vision tasks. Our results demonstrate how hierarchical discriminative learning can improve self-supervised visual representations of biomedical microscopy.

\begin{figure*}[ht!]
    \centering
    \includegraphics[width=\textwidth]{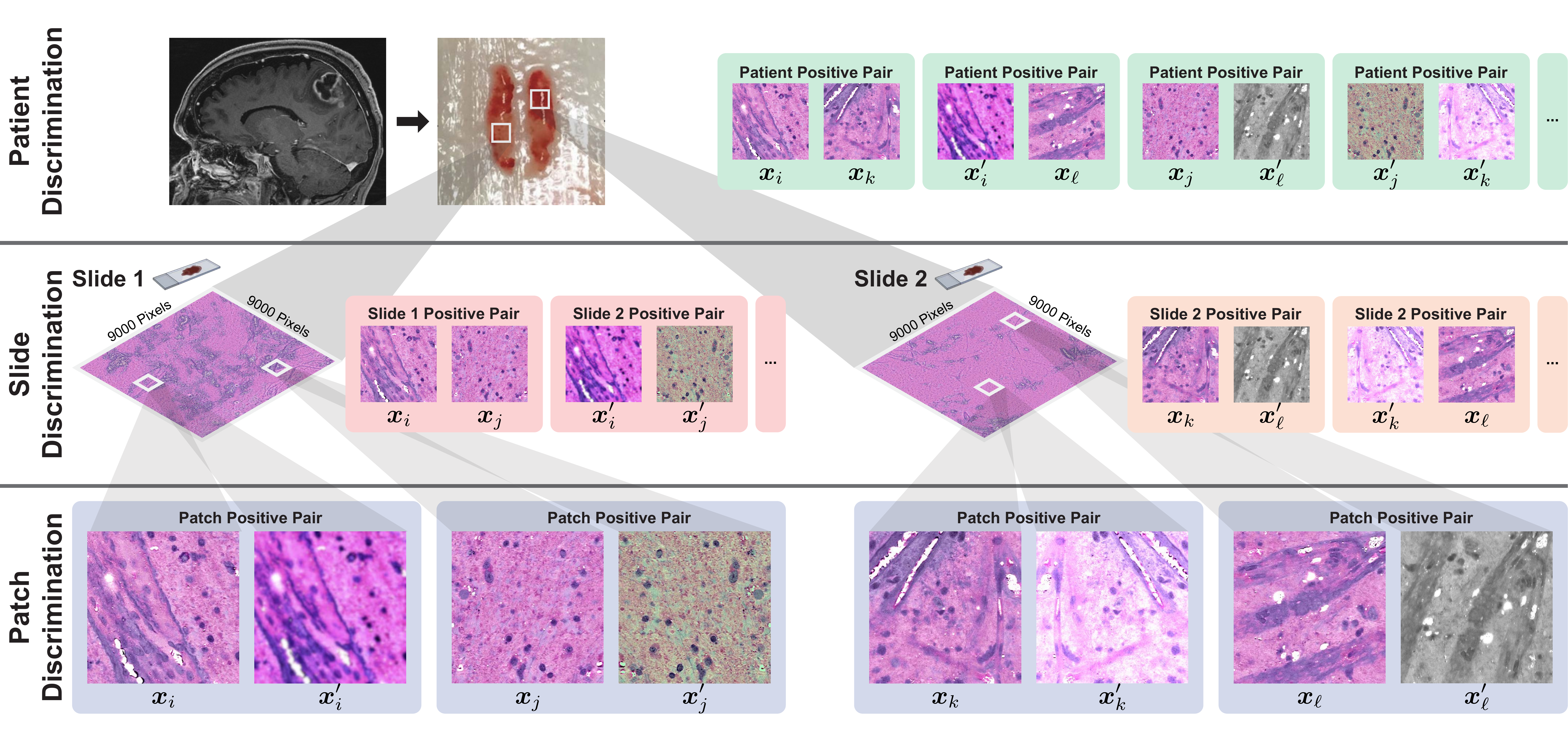}
    \caption{\textbf{HiDisc Overview.} Motivated by the patient-slide-patch data hierarchy of clinical biomedical microscopy, HiDisc defines a patient, slide, and patch discriminative learning objective to improve visual representations. Because WSI and microscopy data are inherently hierarchical, defining a unified hierarchical loss function does not require additional annotations or supervision. Positive patch pairs are defined based on a common ancestry in the data hierarchy. A major advantage of HiDisc is the ability to define positive pairs with\underline{out} the need to sample from or learn a set of strong image augmentations, such as random erasing, shears, color inversion, etc. Because each field-of-view in a WSI is a different view of a patient's underlying cancer diagnosis, HiDisc implicitly learns image features that predict that diagnosis.}
    \label{fig:positive_pairs}
\end{figure*}


\section{Related Work} \label{sec:related_work}
\paragraph{Biomedical microscopy and computational pathology}
Biomedical microscopy refers to a diverse set of microscopy methods used in both clinical medicine and biomedical research. The most common clinical use of biomedical microscopy is light microscopy combined with H\&E staining of clinical tissue specimens, such as tissue biopsies for cancer diagnosis \cite{Louis2021-vd}. The introduction of fast and efficient whole-slide digitization resulted in a rapid increase in the availability of WSIs and accelerated the field of computational pathology \cite{Parwani_undated-lp, Tizhoosh2018-bo}. Computational pathology aims to discover and characterize histopathologic features within biomedical microscopy data for cancer diagnosis, prognostication, and to estimate response to treatment.

The introduction of deep learning to WSI has resulted in clinical-grade computational pathology with diagnostic performance on par with board-certified pathologists \cite{Hou2016-fq, Campanella2019-oc, Hollon2020-ez}. See \cite{Qu2022-qz} for a comprehensive survey of deep learning-based methods in computational pathology. Ilse et al. presented MIL framework using an attention-based global pooling operation for slide-level inference \cite{Ilse2018-ix}. Campanella et al. extended the strategy of a trainable aggregation operation using a recurrent neural network for gigapixel WSI classification tasks \cite{Campanella2019-oc}. Lu et al. updated the attention-based MIL method to allow for better interpretability and efficient weakly supervised  training using transformers \cite{Lu2021-vn}. HiDisc is complementary to any MIL framework and can be used as an effective self-supervised pretraining strategy.

Other biomedical microscopy methods have an increasing role in clinical medicine. Electron \cite{Luse1960-sb}, fluorescence \cite{Sanai2011-ae, Weigert2018-jr, Rivenson2019-hh}, and stimulated Raman scattering microscopy \cite{Freudiger2008-gj, Orringer2017-nn} are a few examples of imaging methods that generate microscopy images used for patient diagnosis. Several studies have applied deep learning-based methods to these modalities for image analysis \cite{Hollon2020-ez, Hollon2020-oj, Jiang2022-ea, Jiang2022-bk, Weigert2018-jr, Rivenson2019-hh}.

\paragraph{Self-supervised pretraining in biomedical microscopy}
Self-supervised pretraining has been used in computational pathology to improve patch-level representation learning \cite{Chen2022-bp, Li2021-qi, Saillard2021-ny, Zhao2020-xm, Qu2022-qz}. Generally, a two-stage approach is used where first an SSL method is applied for patch-level feature learning using instance discrimination, and then the patch-level features are aggregated for slide- or patient-level diagnosis. SSL patch pretraining can reduce the amount of data needed compared to end-to-end MIL training \cite{Qu2022-qz}. Contrastive predictive coding \cite{Lu2019-no}, SimCLR \cite{Li2021-qi}, MoCo \cite{Saillard2021-ny}, VQ-VAE \cite{Chen2022-wn}, and VAE-GAN \cite{Zhao2020-xm} are examples of deep self-supervised visual representation learning methods applied to biomedical microscopy images \cite{Qu2022-qz}. 

Chen et al. presented a study using vision transformers and self-supervised pretraining at different image scales within individual WSIs \cite{Chen2022-bp}. They aim to represent the hierarchical structure of visual histopathologic tokens (e.g., cellular features, supracellular structures, etc.) at varying image resolutions using a transformer pyramid, resulting in a single slide-level representation. Knowledge distillation was used for SSL at each image resolution \cite{Caron2021-lk}. HiDisc is complementary to their method and can be used for SSL at any image resolution or, more generally, field-of-view pretraining for any MIL method for slide-level representations. 

\section{Methods} \label{sec:methods}
\subsection{The Patient-Slide-Patch Hierarchy}
\textit{The motivation for HiDisc is that fields-of-view from clinical WSIs, sampled from within a patient’s tumor, are a diverse set of image examples that capture the same underlying cancer diagnosis}. Our approach focuses on how to use these diverse fields-of-view in the context of the known clinical patient-slide-patch hierarchical structure to improve visual representation learning. Most patients included in public cancer histopathology datasets, including The Cancer Genome Atlas (TCGA) \cite{Cancer_Genome_Atlas_Research_Network2013-au} and OpenSRH \cite{Jiang2022-bk}, contain multiple WSIs as part of their clinical cancer diagnosis. These WSIs may be sampled from different locations in the patient's tumor, or different regions within the same tumor specimen. Both histopathologic and molecular heterogeneity has been well described within human cancers, encouraging clinicians to obtain multiple specimens/samples/views of the patient's tumor \cite{Sottoriva2013-qa}. To leverage the hierarchy shown in Figure \ref{fig:intro}, we create positive pairs at the patch-, slide-, and patient-level to define different discriminative learning tasks with a corresponding increase in visual feature diversity: 
\begin{itemize}
    \item \textbf{Patch discrimination:} Positive pairs are created from different random augmentations of the same patch. This strategy is similar to existing work on SSL via instance discrimination \cite{Chen2020-td, Chen2020-gs, Bardes2022-md}.
    \item \textbf{Slide discrimination:} Positive pairs are created from different augmented patches sampled from the same WSI. These pairs capture local feature diversity within the same specimen. Regional differences in cytologic and histoarchitectural features can be captured at this hierarchical level.
    \item \textbf{Patient discrimination:} Positive pairs are created from different WSIs from the same patient. Patches from different WSIs have the same underlying cancer diagnosis, but can have the greatest degree of feature diversity due to spatially separated tumor specimens. Additionally, diversity in specimen quality, processing, and staining, etc., is captured at this level. 
\end{itemize}

An overview of the hierarchical discrimination tasks is shown in Figure \ref{fig:positive_pairs}.

\begin{table}[b!]
    \centering
    {
    \setlength{\tabcolsep}{5pt}
    \begin{tabular}{c|cc}\hline
        \shortstack{Discrimination\\level} & \shortstack{Number of samples\\ treated as independent } & \shortstack{Number of\\ Positive pairs}\\\hline
        Patch   & $n\cdot n_s \cdot n_p $ & $n_a$\\
        Slide   & $n\cdot n_s$ & $ n_p \cdot n_a$\\
        Patient & $n$ & $n_s\cdot n_p \cdot n_a$\\\hline
    \end{tabular}
    }
    \caption{\textbf{Batch composition at all discrimination levels.} The number of samples in the batch treated as independent and the number of positive pairs at each discrimination level. $n$, number of patients in the batch, $n_s$, number of slides sampled per patient, $n_p$, number of patches sampled per slide, $n_a$, number of augmentations performed on each patch.}\label{tab:bz}
\end{table}

\subsection{Hierarchical Discrimination (HiDisc)}
The formulation of the HiDisc loss function is based on NT-Xent \cite{Chen2020-td} and inspired by \cite{Khosla2020-hy, Zhang2022-zz} for the purpose of multiple positive pairs. The fundamental difference is that no class labels are used during training with a HiDisc loss. HiDisc utilizes the natural hierarchy inherent to biomedical microscopy images to improve self-supervised visual representation learning.

We randomly sample a minibatch of $n$ patients, $n_s$ slides from each patient, $n_p$ patches from each slide, and we augment each patch $n_a$ times. For patients with less than $n_s$ slides, the slides are repeated and sampled. We assume $n_p << $ numbers of patches available for WSI sampling. Note that if a patient has only one WSI, then patient discrimination degenerates to slide discrimination.

\begin{algorithm}[t!]
\caption{HiDisc Pseudocode in PyTorch style}
\label{alg:code}
\definecolor{codeblue}{rgb}{0.0,0.5,0.0}
\definecolor{codekw}{rgb}{0.85, 0.18, 0.50}
\lstset{
  basicstyle=\fontsize{9pt}{9pt}\ttfamily\selectfont,
  columns=fullflexible,
  breaklines=true,
  captionpos=b,
  commentstyle=\color{codeblue},
  keywordstyle=\color{codekw},
}
\begin{lstlisting}[language=python]
# f: backbone + projection mlp
# n: number of patients, batch size
# ns: number of slides sampled per patient
# np: number of patches sampled per slide
# na: number of augmentations per patch
# w: width of each patch
# h: height of each patch
# L: self supervised loss function
# lambdas: weight coefficients for each loss term

for x in loader:
    # Load minibatch x with n patients
    # x.shape is (n, ns, np, na, w, h)
    
    # Forward pass, z.shape is (n * ns * np * na, d)
    z = f(x.reshape(n * ns * np * na, w, h))

    # Compute HiDisc loss for patch,
    # slide and patient level discrimination
    loss_patch   = L(z.reshape(n * ns * np, na, d))
    loss_slide   = L(z.reshape(n * ns, np * na, d))
    loss_patient = L(z.reshape(n, ns * np * na, d))
    loss = dot(lambdas, # lambdas.shape is (1, 3)
        [loss_patch, loss_slide, loss_patient])

    # Back-propagate and update network
    loss.backward()  
    optimizer.step()

\end{lstlisting}
\end{algorithm}

The HiDisc loss consists of the sum of three losses, each of which corresponds to a discrimination task at a different level of the patch-slide-patient hierarchy. Similar to the supervised contrastive learning loss \cite{Khosla2020-hy}, the component HiDisc losses are designed to fit more than one pair of positives within each level of the hierarchy. We define the HiDisc loss at the level $\ell$ to be:
\begin{equation}
    L_\text{HiDisc}^\ell = \sum_{i\in \mathcal{I}}\frac{-1}{|\mathcal{P}_\ell(i)|}
    \sum_{p\in\mathcal{P}_\ell(i)}\log
    \frac{\exp\left(\boldsymbol{z}_i\cdot\boldsymbol{z}_p\right/\tau)}
    {\sum_{a\in\mathcal{A}(i)}\exp\left(\boldsymbol{z}_i\cdot\boldsymbol{z}_a\right/\tau)},
\end{equation}
where $\ell\in\{\text{Patch, Slide, Patient}\}$ is the level of discrimination, and $\mathcal{I}$ is the set of all images in the minibatch. $\mathcal{A}_\ell(i)$ is the set of all images in $\mathcal{I}$ except for the anchor image $i$,
\begin{equation}
    \mathcal{A}(i) = \mathcal{I} \setminus \{i\},
\end{equation}
and $\mathcal{P}_\ell(i)$ is a set of images that are positive pairs of $i$ at the $\ell$-level, 
\begin{equation}
\mathcal{P}_\ell(i) = \{p\in\mathcal{A}_\ell(i): \text{ancestry}_\ell(p) = \text{ancestry}_\ell(i)\},
\end{equation}
where $\text{ancestry}_\ell(\cdot)$ is the $\ell$-level ancestry for an augmented patch in the batch. For example,
patches $\boldsymbol{x}_i$ and $\boldsymbol{x}_j$ from the same patient would have the same patient ancestry, i.e., $\text{ancestry}_\text{Patient}(\boldsymbol{x}_i) = \text{ancestry}_\text{Patient}(\boldsymbol{x}_j)$.

Patch-, slide-, and patient-level HiDisc losses share the same overall contrastive objective, but capture positive pairs at different levels in the hierarchy. Each loss has a different number of positive pairs within a minibatch. Details about this relationship are shown in Table \ref{tab:bz}.

Finally, the complete HiDisc loss is the sum of the patch-, slide-, and patient-level discrimination losses defined above:
\begin{equation}
    \mathcal{L}_\text{HiDisc} = \sum_{\ell \in \{\text{Patch, Slide, Patient}\}}
        \lambda_\ell L_\text{HiDisc}^\ell,
\end{equation}
where $\lambda_\ell$ is a weighting hyperparameter for level $\ell$ in the total loss. The pseudocode in PyTorch-style detailing the training process is shown in Algorithm $\ref{alg:code}$.


\section{Experiments} \label{sec:exp}
We evaluate HiDisc using two different computational histopathology tasks: multiclass histopathologic cancer diagnosis and molecular genetic mutation prediction. We present quantitative classification performance metrics, as well as qualitative evaluation with $t$SNE visualizations \cite{Van_der_Maaten2008-eh} of the learned patch-level representations. 

\subsection{Datasets}
\paragraph{Stimulated Raman histology (SRH)} We validate HiDisc on a multiclass histopathological cancer diagnosis task using an SRH dataset. Stimulated Raman histology is an optical microscopy method that provides rapid, label-free, sub-micron resolution images of unprocessed biological tissues \cite{Freudiger2008-gj, Orringer2017-nn}. The SRH dataset includes specimens from patients who underwent brain tumor biopsy or tumor resection. Patients were consecutively and prospectively enrolled at the University of Michigan for intraoperative SRH imaging, and this study was approved by the Institutional Review Board (HUM00083059). Informed consent was obtained for each patient prior to SRH imaging and approved the use of tumor specimens for research and development. The SRH dataset consists of 852K patches from 3560 slide images from 896 patients with classes consisting of normal brain tissue and 6 different brain tumor diagnoses: high-grade glioma (HGG), low-grade glioma (LGG), meningioma, pituitary adenoma, schwannoma, and metastatic tumor. All slides are divided into 300$\times$300 patches, and they are preprocessed to exclude the empty or non-diagnostic regions using a segmentation model \cite{Hollon2020-ez}.

\paragraph{TCGA diffuse gliomas} We additionally validate HiDisc using WSIs from The Cancer Genome Atlas (TCGA) dataset. We focus on WSIs from brain tumor patients diagnosed with diffuse gliomas, the most common and deadly primary brain tumor \cite{Louis2021-vd}. Molecular genetic mutation classification is used as the evaluation task. The most important genetic mutation that defines lower grade versus high grade diffuse gliomas is isocitrate dehydrogenase-1/2 (IDH) mutational status \cite{Cancer_Genome_Atlas_Research_Network2015-ms}. IDH-mutant tumors are known to have a better prognosis and overall survival (median survival $\sim$10 years) compared to IDH-wildtype tumors ($\sim$1.5 years). The classification task is to predict IDH mutational status using formalin-fixed, paraffin-embedded H\&E-stained WSI images at 20$\times$ magnification from the TCGA dataset. Predicting IDH mutational status from WSIs is currently not feasible for board-certified neuropathologists \cite{Cancer_Genome_Atlas_Research_Network2015-ms, Eckel-Passow2015-oj}; genetic mutation prediction from WSIs could avoid time-consuming and expensive laboratory techniques like genetic sequencing. WSIs are divided into 300$\times$300 pixel fields-of-view, and blank patches are excluded. The rest of the patches are stain normalized using the Macenko algorithm \cite{Macenko2009-bw}. The TCGA data we included consists of a total of 879 patients and 1703 slides, and 11.3M patches.

\subsection{Implementation details}
We train HiDisc using ResNet-50 \cite{He2016-tr} as the backbone feature extractor and a one-layer MLP projection head to project the embedding to 128-dimensional latent space. We use an AdamW \cite{Loshchilov2017-qa} optimizer with a learning rate of 0.001 and a cosine decay scheduler after warmup in the first 10\% of the iterations. For a fair comparison, we control the total number of images in one minibatch ($n\cdot n_a \cdot n_s \cdot n_p$) as 512 and 448 for SRH and TCGA data, respectively. We train HiDisc with patient discrimination by setting $n_a=n_s=n_p=2$, slide discrimination by setting $n_a=n_s=2, n_p=1$, and patch discrimination by setting $n_a = 2, n_s=n_p=1$. The number of patients $n$ sampled from each batch is adjusted accordingly. $\lambda_\ell$ for each level of discriminating loss is set to 1, and temperature $\tau$ is set as 0.7 for all experiments. We define weak augmentation as random horizontal and vertical flipping. The strong augmentations are similar as \cite{Chen2020-td}, including random erasing, color jittering, and affine transformation (for details, see Appendix \ref{sec:app_impl}). We train HiDisc till convergence for both datasets (100K and 60K iterations for SRH and TCGA, respectively) with three random seeds. Training details for all baselines, including SimCLR \cite{Chen2020-td}, SimSiam\cite{Chen2020-gs}, BYOL\cite{Grill2020-at}, VICReg \cite{Bardes2022-md}, and SupCon\cite{Khosla2020-hy}, are similar to HiDisc.

\begin{table*}[ht!]
    \centering
    \setlength{\tabcolsep}{5pt}
\begin{tabular}{c|cc|cc||cc|cc}\hline
    \multirow{2}{*}{} & \mulccol{SRH - Patch} & \multicolumn{2}{c||}{SRH - Patient} & \multicolumn{2}{c}{TCGA - Patch} & \mulccol{TCGA - Patient}\\\cline{2-9}
                  & Accuracy & MCA  & Accuracy  & MCA &  Accuracy & AUROC  & Accuracy  & AUROC \\\hline
SimCLR\cite{Chen2020-td}   & 81.0 (0.1) & 73.9 (0.2) & 83.1 (0.7) & 78.4 (0.6) &                   77.8 (0.0) & 85.2 (0.1) & 80.7 (0.6) & 88.9 (0.3)\\
SimSiam\cite{Chen2020-gs}  & 80.3 (1.9) & 73.6 (2.7) & 82.3 (1.7) & 77.0 (4.0) &                   68.4 (0.3) & 74.1 (0.3) & 76.6 (0.6) & 82.4 (0.3)\\
BYOL\cite{Grill2020-at}    & 83.5 (0.1) & 78.2 (0.2) & 84.8 (1.0) & 82.7 (1.0) &                   80.0 (0.1) & 87.5 (0.1) & 83.1 (0.6) & 89.8 (0.2)\\
VICReg\cite{Bardes2022-md} & 82.1 (0.3) & 76.0 (0.4) & 82.1 (0.7) & 78.7 (1.9) &                   75.5 (0.1) & 82.9 (0.1) & 77.0 (0.5) & 86.0 (0.3)\\\hline
HiDisc-Patch               & 80.8 (0.0) & 73.5 (0.1) & 82.6 (0.3) & 77.9 (0.3) &                   77.2 (0.1) & 84.7 (0.1) & 81.0 (0.5) & 88.1 (0.2)\\
HiDisc-Slide               & 86.9 (0.2) & \tbf{83.2 (0.2)} & \tbf{87.6 (0.5)} & \tbf{87.0 (1.4)} &       82.7 (0.2) & 89.3 (0.2) & \tbf{84.3 (0.3)} & \tbf{92.3 (0.3)}\\
HiDisc-Patient             & \tbf{87.4 (0.1)} & \tbf{83.5 (0.2)} & \tbf{87.9 (0.5)} & \tbf{86.4 (0.6)} & \tbf{83.1 (0.1)} & \tbf{90.1 (0.1)} & \tbf{83.6 (0.3)} & \tbf{91.8 (0.2)}\\\hline
Supervised                 & 88.9 (0.3) & 86.3 (0.3) & 88.5 (0.5) & 89.1 (0.5) &                   85.1 (0.3) & 91.7 (0.2) & 88.3 (0.4) & 95.2 (0.2)\\\hline
\end{tabular}
    
    \caption{\textbf{Main results.} We compare HiDisc to state-of-the-art SSL. Supervised refers to models trained with supervision from class labels. Standard deviations are reported in parentheses. HiDisc-Slide and HiDisc-Patient outperform all baseline methods in all metrics for both tasks. Our HiDisc benchmark outperforms previously reported fully \textit{supervised} baselines from the existing literature on the same genetic mutation classification task on TCGA \cite{Liu2020-cl}. MCA, mean class accuracy, AUROC, area under the receiver operating characteristic curve.} \label{tab:main}
\end{table*}

\begin{figure*}[ht!]
    \centering
    \includegraphics[width=\textwidth]{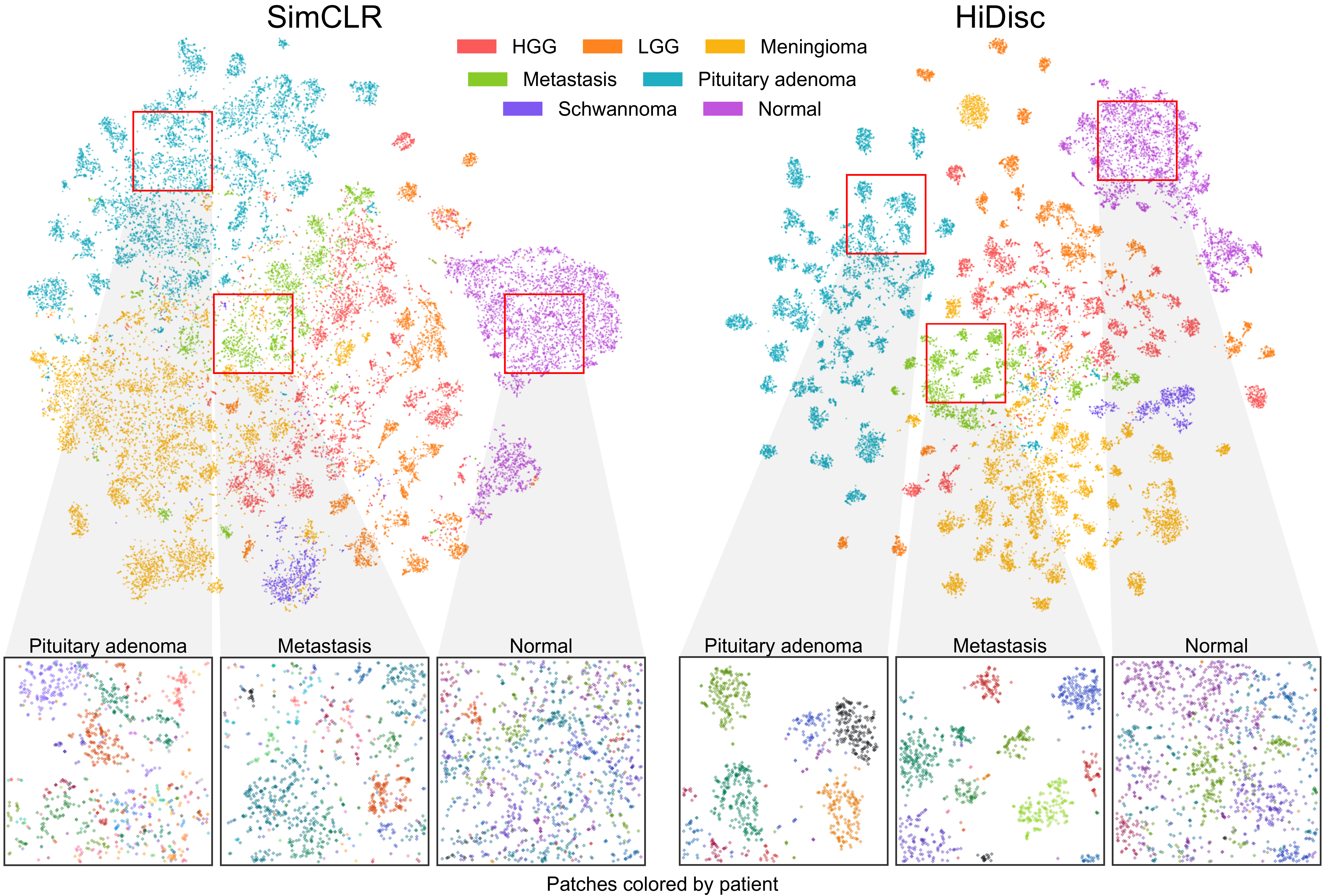}
    \caption{\textbf{Visualization of learned SRH representations using SimCLR and HiDisc.} 
    \textbf{Top.} Randomly sampled patch representations are visualized after SimCLR versus HiDisc pretraining using $t$SNE \cite{Van_der_Maaten2008-eh}. Representations are colored based on brain tumor \textit{diagnosis}. HiDisc qualitatively achieves higher quality feature learning and class separation compared to SimCLR. Expectedly, HiDisc shows within-diagnosis clustering that corresponds to patient discrimination. \textbf{Bottom.} Magnified cropped regions of the above visualizations show subclusters that correspond to individual patients. Patch representations in magnified crops are colored according to \textit{patient} membership. We see patient discrimination within the different tumor diagnoses. Importantly, we do not see patient discrimination within normal brain tissue because there are minimal-to-no differentiating microscopic features between patients. This demonstrates that in the absence of discriminative features at the slide- or patient-level, HiDisc can achieve good feature learning using patch discrimination without overfitting the other discrimination tasks. HGG, high grade glioma; LGG, low grade glioma; Normal, normal brain tissue.}
    \label{fig:tsne_big}
\end{figure*}

\subsection{Evaluation protocols}
\paragraph{$\boldsymbol{k}$NN evaluation}
Standard protocols to evaluate self-supervised representation learning include linear and fine-tuning evaluation. However, both methods are sensitive to hyperparameters, such as learning rate\cite{Caron2021-lk}. Therefore, we use the $k$ nearest neighbor ($k$NN) classifier for quantitative evaluation. We freeze the pretrained backbone to compute the representation vectors for both training and testing data. The nearest neighbor classifier is then used to match each patch in the testing dataset to the most similar $k$ patch representation vectors in the training dataset based on their cosine similarity.  It also outputs a prediction score measured by the cosine similarity between a test image and its $k$ nearest neighbors. Due to the size of the TCGA dataset, we randomly sample 400 patches from each slide for evaluation across three different random seeds. Using the $k$NN classifier, we can compute accuracy (ACC), mean class accuracy (MCA), and area under receiver operating characteristic curve (AUROC) for patch metrics. We use MCA for SRH dataset because it is a multiclass classification problem and the classes are imbalanced. The AUROC is used for the TCGA dataset since it is a balanced binary classification task.

\paragraph{Slide and patient metrics}
In contrast to patch-level evaluation, slide and patient predictions are more practical for cancer diagnosis and other clinical uses\cite{Campanella2019-oc}. After getting patch prediction scores by $k$NN evaluation, we use average pooling over the scores within each WSI or patient to obtain an aggregated prediction score. Compared to most MIL methods, this non-parametric method directly evaluates representation learning without additional training.

\begin{table*}[ht!]
    \centering
    \setlength{\tabcolsep}{5pt}
    \begin{tabular}{c|cc|cc||cc|cc}\hline
        \multirow{2}{*}{} & \mulccol{SRH - Patch} & \multicolumn{2}{c||}{SRH - Patient} & \multicolumn{2}{c}{TCGA - Patch} & \mulccol{TCGA - Patient}\\\cline{2-9}
                      & Accuracy & MCA  & Accuracy  & MCA &  Accuracy & AUROC  & Accuracy  & AUROC \\\hline
SimCLR\cite{Chen2020-td}        & 31.5 (2.3) & 23.1 (1.9) & 40.2 (6.9) & 28.9 (4.5) &                       57.1 (1.1) & 58.4 (2.2) & 58.1 (1.1) & 72.8 (3.2)\\\hline
HiDisc-Patch                    & 31.3 (0.6) & 22.2 (0.5) & 47.4 (2.1) & 33.1 (1.6) &                       59.0 (0.8) & 61.5 (1.3) & 61.2 (4.2) & 75.8 (2.5)\\
HiDisc-Slide                    & 82.8 (0.2) & 77.4 (0.3) & \tbf{84.2 (0.5)} & \tbf{82.3 (0.4)} &           79.6 (0.1) & 86.3 (0.2) & 77.7 (0.4) & 85.3 (0.3)\\
HiDisc-Patient                  & \tbf{84.9 (0.2)} & \tbf{78.9 (0.1)} & \tbf{84.7 (0.5)} & \tbf{80.9 (1.4)} &     \tbf{82.9 (0.2)} & \tbf{89.6 (0.2)} & \tbf{82.3 (0.3)} & \tbf{90.3 (0.3)}\\\hline
SupCon\cite{Khosla2020-hy}      & 90.0 (0.2) & 87.4 (0.3) & 90.0 (0.5) & 90.3 (0.4) &                       85.4 (0.4) & 92.0 (0.2) & 88.4 (0.8) & 95.2 (0.4)\\\hline
    \end{tabular}
    
    \caption{\textbf{Results for weak augmentation.} Comparison of HiDisc with the SimCLR and SupCon on SRH and TCGA datasets with weak augmentation (random flipping). Both SimCLR and HiDisc-Patch patch collapse since weak augmentation will make the pretext task trivial. SupCon, on the other hand, is not sensitive to data augmentations since its positive pairs are defined by class labels. HiDisc-Slide and HiDisc-Patient are only slightly affected by the augmentation and achieve performance close to the supervised method.}
    \label{tab:weak_aug}
\end{table*}


\section{Results} \label{sec:results}
\subsection{Quantitative metrics} In this section, we evaluate the representations learned by self-supervised HiDisc pretraining using the training and evaluation protocols described in Section \ref{sec:exp}. We compare HiDisc and SSL baselines on the testing set of both datasets in Table \ref{tab:main}. Since HiDisc-Patch is most similar to SimCLR, we observe similar performances. After we add slide and patient discrimination in HiDisc-Slide and HiDisc-Patient, we observe a significant boost in patch accuracy (+6.1 and +6.6 on SRH, +5.5 and +5.9 on TCGA), and a similar increase in other performance metrics as well. Among all methods, HiDisc-Patient has the best performance and outperforms the best baseline, BYOL, with an improvement of 3.9\% and 3.1\% in classification accuracy for SRH and TCGA, respectively. The GPU wall time needed to train BYOL is roughly 1.5x longer than HiDisc because it requires updating the weights of the target network using exponential moving average. Appendix \ref{sec:app_results} shows additional model evaluation metrics.

\subsection{Qualitative evaluation} 
\begin{figure}[b!]
    \centering
    \includegraphics[width=\columnwidth]{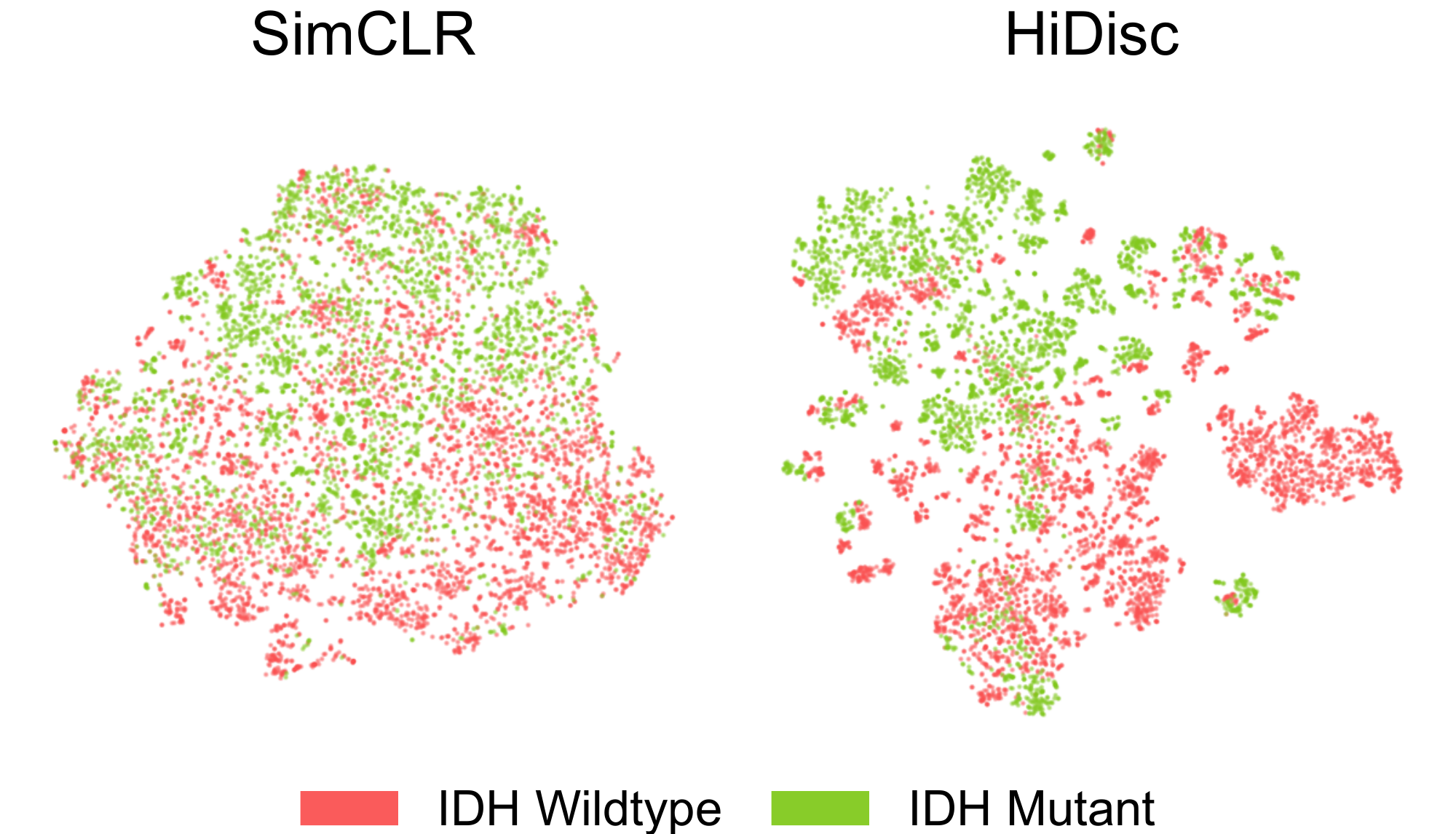}
    \caption{\textbf{Visualization of learned TCGA representations using SimCLR and HiDisc.} We randomly sample patches from the validation set, and visualize these representations using $t$SNE \cite{Van_der_Maaten2008-eh}. Representations on the plots are colored by IDH mutational status. Qualitatively, we can observe that HiDisc forms better representations compared to SimCLR, with clusters within each mutation that corresponds to patient membership. This observation is consistent with the visualizations for the SRH dataset in Figure \ref{fig:tsne_big}.
    \label{fig:tcga_tsne}}
\end{figure}

We also qualitatively evaluate the learned patch representations with $t$SNE \cite{Van_der_Maaten2008-eh} colored by class and patient label for both SimCLR and HiDisc. Figures \ref{fig:tsne_big} and \ref{fig:tcga_tsne} show the learned representations for the SRH and TCGA datasets, respectively. Here, we randomly sample patches from the validation set and plot them by class membership (tumor or molecular subtype). We observe that HiDisc learns better representations for both classification tasks, with more discernible clusters for each class. We also observe better patient clusters within each tumor class in the representations learned by HiDisc. Furthermore, patient clusters are not observed in normal brain tissues for both SimCLR and HiDisc-Patch, as there are minimal differentiating microscopic features between patients.

\subsection{Ablation Studies}
\begin{figure}[t!]
    \centering
    \includegraphics[width=\columnwidth]{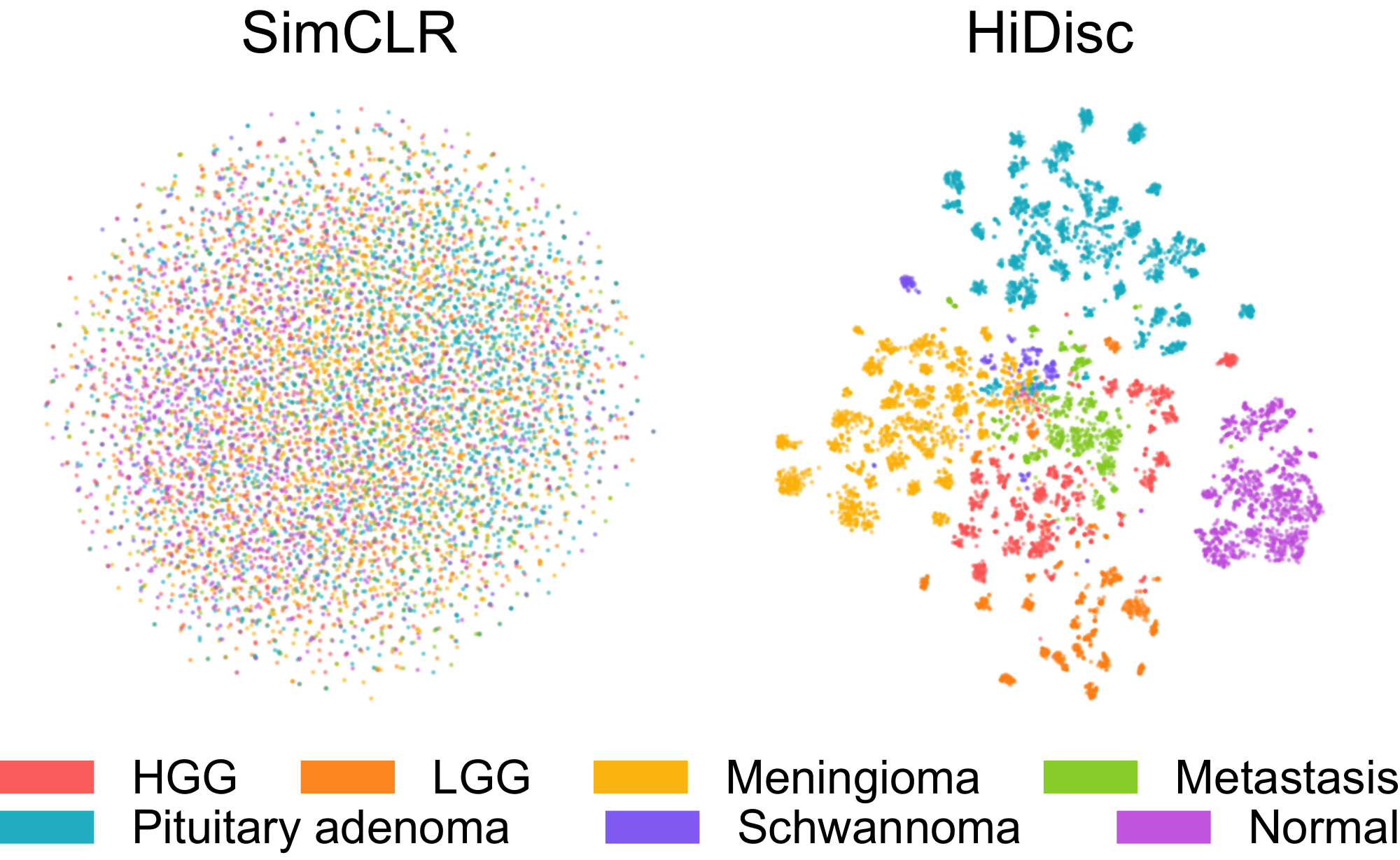}
    \caption{\textbf{Hierarchical self-supervised discriminative learning with\underline{out} strong data augmentations.} Randomly sampled patch representations are shown after SimCLR and HiDisc-Patient pretraining without the use of strong, domain-agnostic data augmentations on SRH dataset. HiDisc achieves high-quality representation learning, while SimCLR is unable to learn semantically meaningful features via instance discrimination alone. HGG, high grade glioma; LGG, low grade glioma; Normal, normal brain tissue}
    \label{fig:tsne_noaug}
\end{figure}

\paragraph{Weak Augmentation} We demonstrate that HiDisc is capable of preserving excellent performance without the use of strong, domain-agnostic data augmentations as shown in Table \ref{tab:weak_aug}. SimCLR suffers from dimensional collapse with weak augmentations \cite{Jing2021-wn}. Similar to SimCLR, HiDisc-Patch collapses because it is limited to the diversity from augmentations alone. HiDisc-Slide and HiDisc-Patient performance remain high across both datasets and tasks. HiDisc-Patient outperforms HiDisc-Slide, especially when evaluating at the patch level. We hypothesize that this is a result of additional diversity between positive pairs contributed by patient-level discrimination. We also provide supervised contrastive learning (SupCon) baselines \cite{Khosla2020-hy} as an upper performance bound. Figure \ref{fig:tsne_noaug} shows SimCLR fails to learn semantically meaningful features while HiDisc achieves high-quality representations. Overall, we observe that HiDisc, especially HiDisc-Patient, performs well regardless of whether strong augmentations are used.

\paragraph{Other Ablation studies} Additional experiments are included in Appendix \ref{sec:app_supp}. We perform ablation studies on the \textbf{weighting factor} $\lambda_\ell$ for each level of discrimination. We also train HiDisc with different \textbf{number of iterations}, \textbf{learning rate} and \textbf{batch size}. Some settings may marginally improve model performance but cost a significant amount of computation resources.

\section{Conclusion} \label{sec:conclusion}
We present HiDisc, a unified, hierarchical, self-supervised representation learning method for biomedical microscopy. HiDisc is able to outperform other state-of-the-art SSL methods for visual representation learning. The performance increase is driven by leveraging the inherent patient-slide-patch hierarchy of clinical WSIs. The inherent data hierarchy provides image diversity by defining positive patch pairs based on a common ancestry in the hierarchy and does not require strong, domain-agnostic augmentations. By combining patch, slide, and patient discrimination into a single learning objective, HiDisc implicitly learns image features of the patient's underlying diagnosis without the need for patient-level annotations or supervision. 

\paragraph{Limitations} Like other WSI classification methods, HiDisc representation learning is currently limited to single-resolution fields-of-view that are arbitrarily defined. Expanding HiDisc to include multiple image resolutions could improve its ability to capture multiscale image features of the patient's underlying diagnosis. Also, we have limited the evaluation to a contrastive learning framework and HiDisc can also be evaluated using other siamese learning frameworks \cite{Bardes2022-md, Chen2020-gs, Caron2021-lk}. 

\paragraph{Broader Impacts} We have limited this investigation to biomedical microscopy and WSIs. However, many imaging medical modalities, such as magnetic resonance imaging and fundoscopy \cite{Gulshan2016-fr}, have a clinical hierarchical data structure that could benefit from a similar hierarchical representation learning framework. We hope that hierarchical discriminative learning will extend beyond microscopy to other medical and non-medical imaging domains. 


\section*{Acknowledgements and Competing Interests}\label{sec:ack}
We would like to thank Karen Eddy, Lin Wang, Andrea Marshall, Eric Landgraf, and Alexander Gedeon for their support in data collection and programming. 

This work is supported in part by grants NIH/NIGMS T32GM141746, NIH K12 NS080223, NIH R01CA226527, Cook Family Brain Tumor Research Fund, Mark Trauner Brain Research Fund: Zenkel Family Foundation, Ian’s Friends Foundation, and the Investigators Awards grant program of Precision Health at the University of Michigan.

C.W.F. is an employee and shareholder of Invenio Imaging, Inc., a company developing SRH microscopy systems. D.A.O. is an advisor and shareholder of Invenio Imaging, Inc, and T.C.H. is a shareholder of Invenio Imaging, Inc.

{\small
\bibliographystyle{include/cvpr/ieee_fullname}
\bibliography{hidisc.bib}
}


\appendix\clearpage
\definecolor{lg}{HTML}{ddddde}
\section{Experimentation Details} \label{sec:app_impl}
\subsection{Dataset statistics}
We perform our experiments on an internal stimulated Raman histology (SRH) dataset and light microscopy images from the publicly available TCGA dataset. Both datasets are split randomly into training and evaluation sets at the patient-level.

\paragraph{SRH dataset.} The SRH dataset is collected using a commercially available NIO microscope (Invenio Imaging, inc, CA), following the protocol and descriptions in \cite{Orringer2017-nn}. Our SRH dataset consists of 852K patches, 3560 whole-slide images from 896 patients in 7 different classes. The detailed training and evaluation set breakdown is listed in \ref{tab:srh_stats}.

\begin{table}[h!]
    \centering{\setlength{\tabcolsep}{2pt}
    \begin{tabular}{c|cccccc}\hline
        \shortstack{Tumor\\class} & \shortstack{\# Train\\Patients} & \shortstack{\# Train\\Slides} & \shortstack{\# Train\\Patches} & \shortstack{\# Eval\\Patients} & \shortstack{\# Eval\\Slides} & \shortstack{\# Eval\\Patches}\\\hline
HGG    & 149 & 541 & 139K & 36 & 132 & 30K\\
LGG    & 93  & 313 & 139K & 24 & 107 & 22K\\
Mening & 154 & 533 & 130K & 47 & 204 & 22K\\
Met    & 93  & 331 &  64K & 24 & 114 & 17K\\
Pit    & 145 & 527 & 137K & 46 & 194 & 30K\\
Schwan & 15  & 47  &  10K & 5  & 22  &  5K\\
Normal & 99  & 343 &  82K & 27 & 152 & 30K\\\hline
    \end{tabular}}
    \caption{SRH dataset number of patients, slides, and patches breakdown of each class. HGG, high grade glioma; LGG, low grade glioma; mening, meningioma; met, metastasis; pit, pituitary adenoma; schwan, schwannoma; normal, normal brain tissue.}
    \label{tab:srh_stats}
\end{table}

\paragraph{TCGA dataset.} TCGA is a large-scale, multicenter consortium that includes biospecimens from 33 cancer types. We focus on brain tumor specimens from the TCGA-GBM and TCGA-LGG studies to evaluate HiDisc for diffuse glioma molecular genetic classification. The brain specimens in the TCGA dataset contain over 10 million patches from 29 institutions. Detailed dataset breakdown for each class is shown in Table \ref{tab:tcga_he_stats}.
\begin{table}[h!]
    \centering{\setlength{\tabcolsep}{2pt}
    \begin{tabular}{c|cccccc}\hline
        \shortstack{IDH\\status} & \shortstack{\# Train\\Patients} & \shortstack{\# Train\\Slides} & \shortstack{\# Train\\Patches} & \shortstack{\# Eval\\Patients} & \shortstack{\# Eval\\Slides} & \shortstack{\# Eval\\Patches}\\\hline
Wt   & 367 & 732 & 4.7M & 92 & 191 & 1.2M\\
Mut  & 336 & 626 & 4.4M & 84 & 154 & 1.1M \\\hline
    \end{tabular}}
    \caption{TCGA dataset number of patients, slides, and patches breakdown of each class. IDH, isocitrate dehydrogenase (IDH-1/2); wt, wildtype; mut, mutant.}
    \label{tab:tcga_he_stats}
\end{table}

\subsection{Implementation}
\paragraph{Augmentations.}
The strong augmentations we use in HiDisc training follow \cite{Chen2020-td}, and include the following augmentations applied sequentially, with a probability of 0.3 for each augmentation. Default PyTorch parameters are used, unless otherwise specified.
\begin{itemize}[noitemsep, nolistsep, topsep=0pt]
    \item Random horizontal and vertical flip;
    \item Gaussian noise;
    \item Color jittering;
    \item Random autocontrast;
    \item Random solarize with threshold 0.2;
    \item Random adjust sharpness with sharpness factor 2;
    \item Gaussian blur with kernel size 5 and sigma 1;
    \item Random erasing;
    \item Random affine transformation with max 10 degrees rotation and 10-30\% image translation;
    \item Random resized crop.
\end{itemize}

\begin{figure}[t!]
    \centering
    \includegraphics[width=\columnwidth]{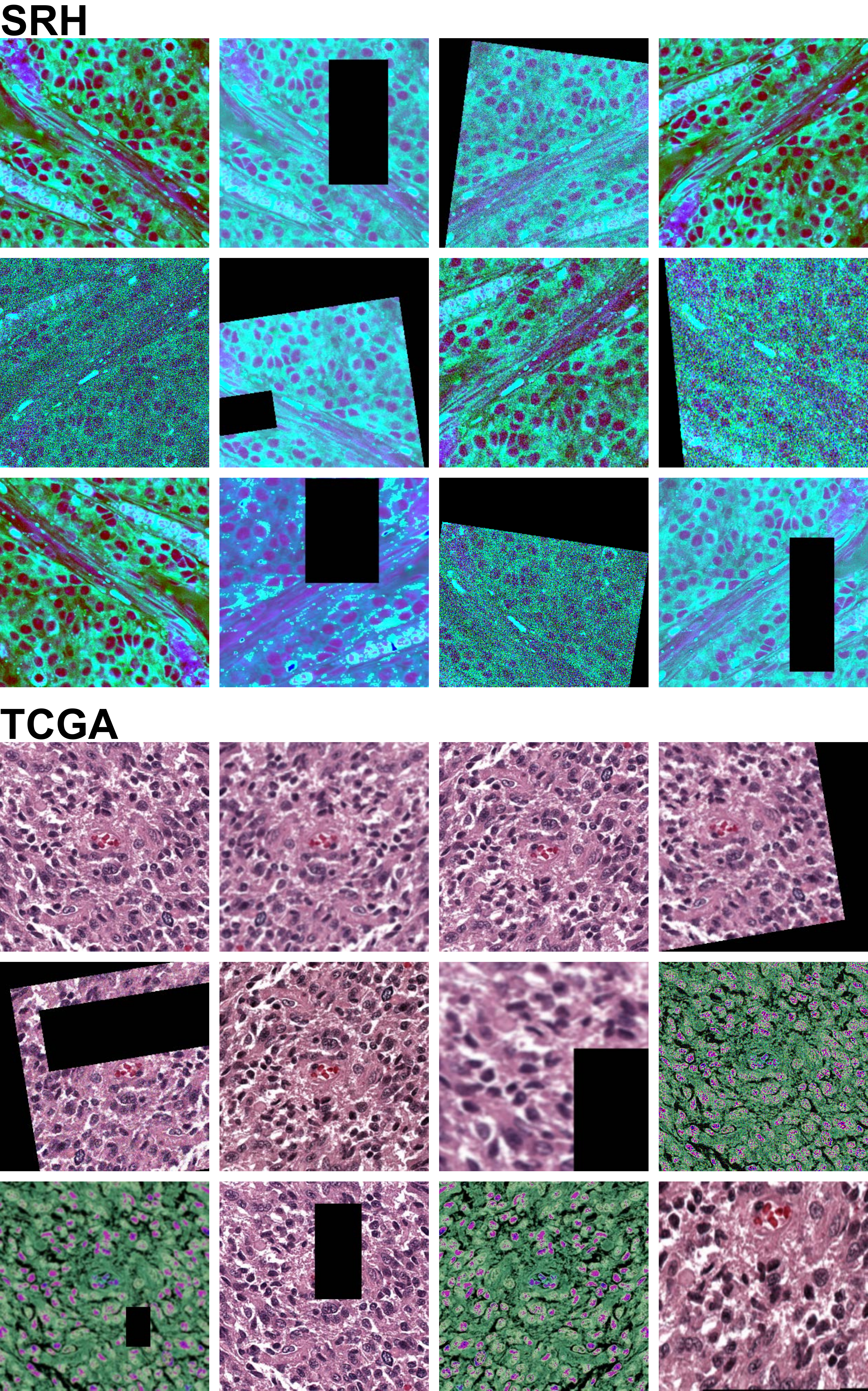}
    \caption{\textbf{SRH and TCGA augmentation panel.} Panels demonstrate examples of augmented image patches. Top left in each panel is the original, and the other patches are generated by randomly sampling from our set of strong augmentations. While these augmentations can help to regularize training and improve generalizability, they have been shown to decrease performance on some histopathology classification tasks.}
    \label{fig:aug_panel}
\end{figure}

Figure \ref{fig:aug_panel} demonstrates the random strong augmentations for the SRH and the TCGA datasets, respectively.

\paragraph{Filtering and Preprocessing.} Following prior work, we divide all whole-slide images into 300$\times$300 patches. We use a previously trained tumor segmentation model \cite{Hollon2020-ez} to filter out blank and non-diagnostic patches from the SRH dataset. For the TCGA dataset, a heuristic algorithm based on the standard deviation of pixel values is used. TCGA patches are then normalized using the Macenko algorithm \cite{Macenko2009-bw}. 

\section{Extended Experimentation Metrics} \label{sec:app_results}
\begin{figure*}[ht!]
    \centering
    \includegraphics[width=\textwidth]{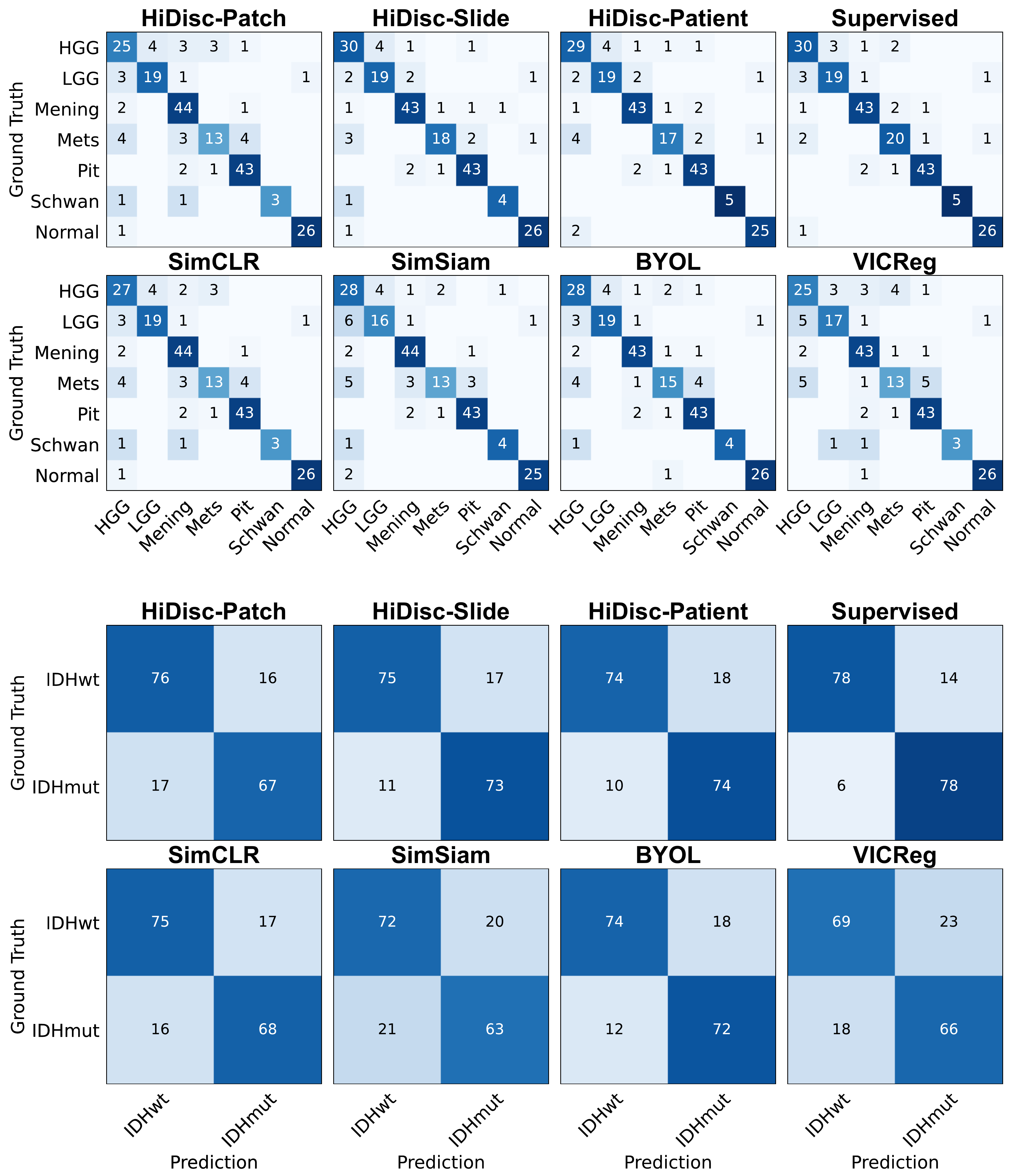}
    \caption{\textbf{Patient-level confusion matrices for main experiments.} These confusion matrices correspond to experiments reported in table \ref{tab:main}. We can observe that HiDisc-Slide and HiDisc-Patient achieve superior performance compared to existing methods. HGG, high grade glioma, LGG, low grade glioma, mening, meningioma, mets, metastasis, pit, pituitary adenoma, schwan, schwannoma, normal, normal brain tissue, IDHwt, IDH wildtype, IDHmut, IDH mutant.}
    \label{fig:main_conf}
\end{figure*}

We present our main results in Table \ref{tab:main}. In addition to patch and patient-level evaluation, we compute slide-level metrics by aggregating the prediction on each whole slide image. For both SRH and TCGA experiments, we add area under the precision-recall curve (AUPRC), and for TCGA experiments, we also include sensitivity and specificity. The extended main results are in Tables \ref{tab:main_srh_extended} and \ref{tab:main_tcga_extended} for SRH and TCGA, respectively. The metrics reported in these tables are consistent with Table \ref{tab:main}, with slide and patient discrimination in HiDisc outperforming existing contrastive learning baselines across multiple different metrics and different levels. Confusion matrices are included in Figure \ref{fig:main_conf}.

\section{Additional Ablation Studies}  \label{sec:app_supp}
\subsection{Weak augmentations}
\begin{figure*}[ht!]
    \centering
    \includegraphics[width=\textwidth]{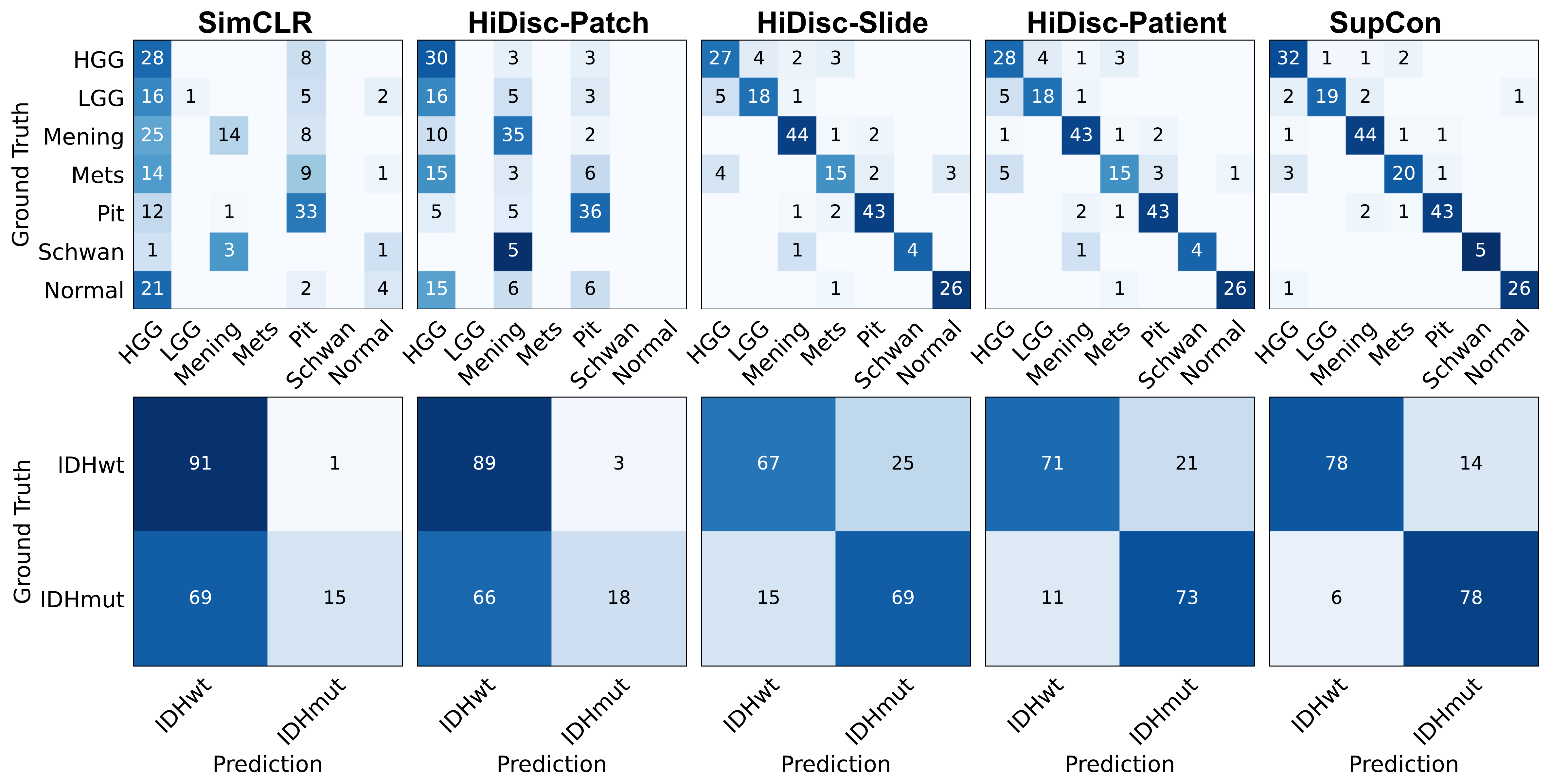}
    \caption{\textbf{Patient-level confusion matrices for experiments with weak augmentations.} These confusion matrices correspond to experiments reported in table \ref{tab:weak_aug}. We can observe that HiDisc-Slide and HiDisc-Patient achieve superior performance compared to existing methods. As expected, patch discrimination methods, such as SimCLR and HiDisc-Patch, collapse because they fail to provide a meaningful pretext task to learn a good representation with weak augmentation. HGG, high grade glioma, LGG, low grade glioma, mening, meningioma, mets, metastasis, pit, pituitary adenoma, schwan, schwannoma, normal, normal brain tissue, IDHwt, IDH wildtype, IDHmut, IDH mutant.}
    \label{fig:weak_conf}
\end{figure*}

We also report the same additional metrics for our experiments with weak augmentation in Table \ref{tab:weak_aug}. The extended results are in Tables \ref{tab:weak_srh_extended} and \ref{tab:weak_tcga_extended} for SRH and TCGA datasets, respectively, and the confusion matrices are reported in Figure \ref{fig:weak_conf}. While we observe a slight reduction in accuracy metrics at all levels for HiDisc, models trained using only instance discrimination, like SimCLR, collapse as expected because it fails to provide a meaningful pretext task to learn a good representation with weak augmentation. This can also be observed in the confusion matrices, where predictions from SimCLR and HiDisc-Patch are overwhelmingly in the majority classes.

\subsection{Weighting factor lambdas}
We conduct ablation on the effect of weighted factor $\lambda$ at different level of discrimination. 
All experiments use the HiDisc-Patient with strong augmentations. HiDisc-Patient utilizes all levels of discrimination but is weighted by different $\lambda$ values. For SRH experiments, we set one of the $\lambda$ values to 0 or 5, and for TCGA experiments, we set one of the $\lambda$ values to 0, 2, or 5, as well as setting two of the $\lambda$ values to 0. The metrics are reported in Table \ref{tab:srh_tune_lambda} and \ref{tab:tcga_tune_lambda} for SRH and TCGA, respectively. HiDisc is relatively robust to changes in $\lambda_\text{Patient}$ and $\lambda_\text{Slide}$ values, as slide- and patient-level discrimination are complementary to each other. As expected, when $\lambda_\text{Patient}=\lambda_\text{Slide}=0$, we observe a significant reduction in model performance because only patch discrimination is used to supervise model training. Interestingly, we can see a slight performance drop when $\lambda_\text{Patch}$ is amplified, and removing patch discrimination slightly boosted performance in SRH.

\subsection{Learning rates} 
\begin{figure}[h!]
    \centering
    \includegraphics[width=\columnwidth]{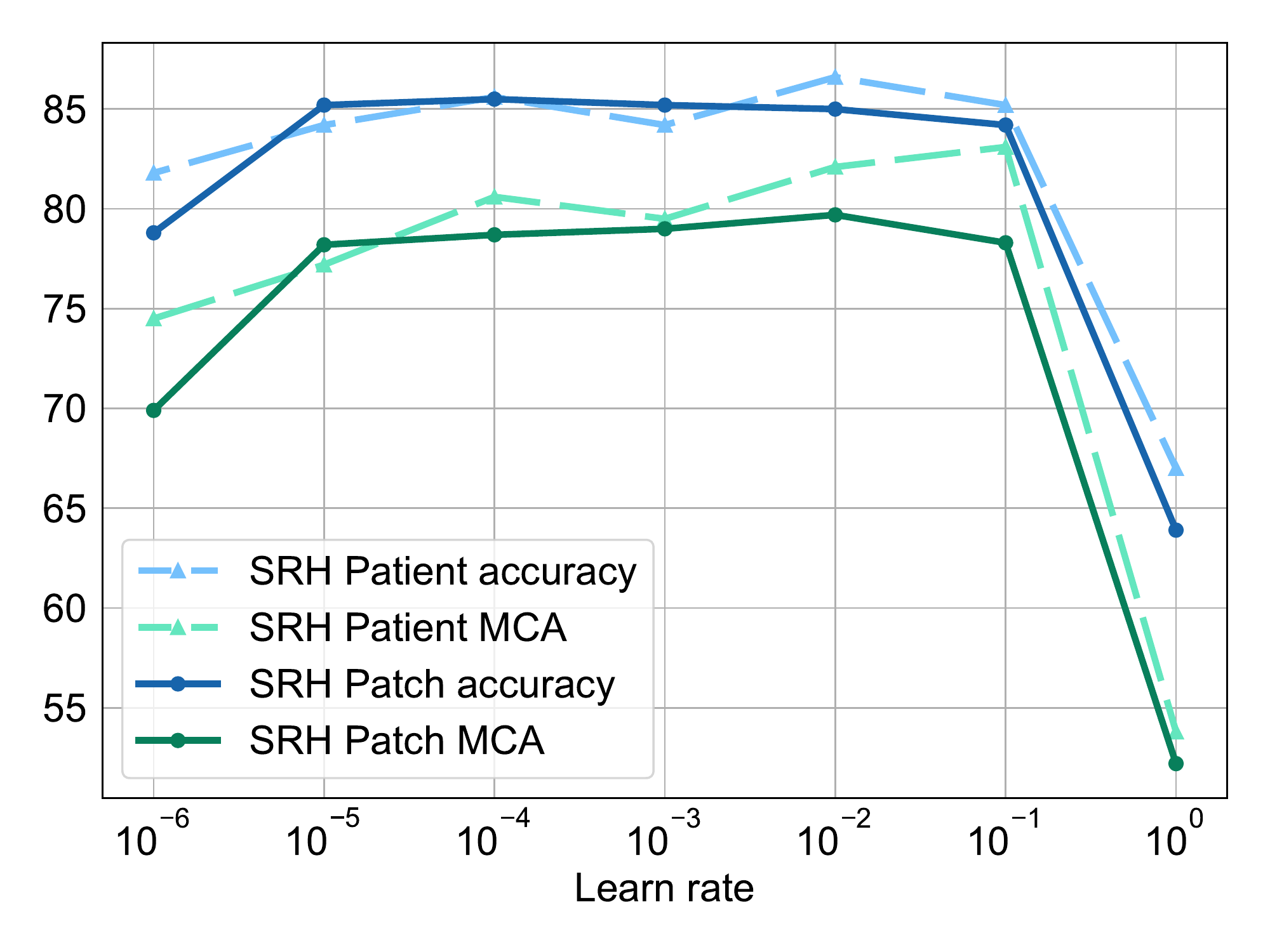}
    \caption{\textbf{Learning rate ablation.} HiDisc-Patch models are trained with batch size 512. We choose a wide range of learning rates and it shows HiDisc performs robustly from $\mathrm{10^{-1}}$ to $\mathrm{10^{-5}}$. MCA, mean class accuracy.}
    \label{fig:lr}
\end{figure}

We evaluate model performance with different learning rates, and the model performances for SRH and TCGA datasets are reported in Table \ref{tab:srh_tune_lr} and \ref{tab:tcga_tune_lr}, respectively. The HiDisc performance on the SRH dataset vs learning rate is also summarized in Figure \ref{fig:lr}. We can observe that HiDisc training is robust to variation in learning rate, achieving good performance on the SRH dataset from $\mathrm{10^{-1}}$ to $\mathrm{10^{-5}}$.

\subsection{Batch sizes} 

Literature for early contrastive learning algorithms such as SimCLR \cite{Chen2020-td} shows benefits from training with larger batch size. We perform ablation studies to investigate the effect of batch size on HiDisc training. Due to the computation resources limit, we are only able to ablate batch size on 512 and 1024 for SRH dataset. The batch size is defined here as the total number of images including augmentation in one batch. As shown in Table \ref{tab:srh_tune_bs}, we do not observe a significant benefit of using a larger batch size. 

\subsection{Iterations}
\begin{figure}[h!]
    \centering
    \includegraphics[width=\columnwidth]{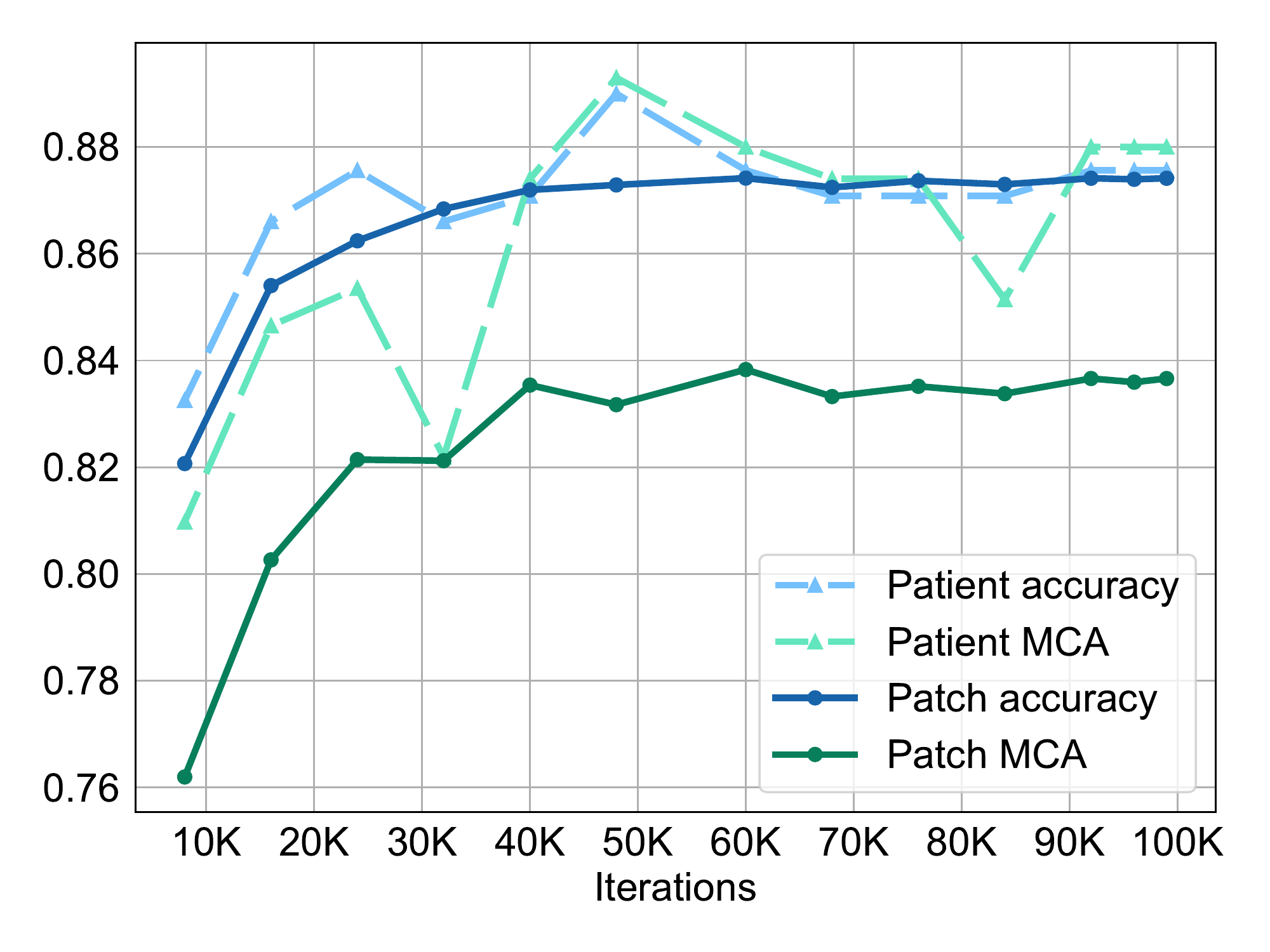}
    \caption{\textbf{Iterations ablation.} We empirically show the convergence of HiDisc until 100K iterations of training. HiDisc achieves a performance plateau after 40K iterations. MCA, mean class accuracy.}
    \label{fig:iter}
\end{figure}

We present the training curve of HiDisc-Patient on the SRH dataset as an example in Figure \ref{fig:iter} (showing validation set metrics). This experiment uses batch size 512, learning rate as $\mathrm{10^{-3}}$ and strong augmentation. We can observe HiDisc does not need long training time and achieve a performance plateau after 40K iterations of training.

\clearpage

\begin{table*}[ht!]
    \centering
    \setlength{\tabcolsep}{4pt}
    \begin{tabular}{c|ccc||>{\columncolor{lg}}c>{\columncolor{lg}}c>{\columncolor{lg}}c||ccc}\hline
    & \multicolumn{3}{c||}{\tbf{Patch} Level Metrics} & \multicolumn{3}{>{\columncolor{lg}}c||}{\tbf{Slide} Level Metrics} & \multicolumn{3}{c}{\tbf{Patient} Level Metrics}\\
    \hhline{>{\arrayrulecolor{white}}-*{3}{>{\arrayrulecolor{black}}|-}*{3}{>{\arrayrulecolor{black}}|-}*{3}{>{\arrayrulecolor{black}}|-}|}
        Method & Accuracy & MCA  & AUPRC& Accuracy & MCA  & AUPRC& Accuracy & MCA  & AUPRC\\\hline
SimCLR         & 81.0 (0.1) & 73.9 (0.2) & 81.5 (0.2) & 82.1 (0.3) & 76.1 (0.3) & 87.8 (0.2) & 83.1 (0.7) & 78.4 (0.6) & 87.8 (0.2)\\
SimSiam        & 80.3 (1.9) & 73.6 (2.7) & 79.5 (3.7) & 81.4 (1.8) & 75.1 (2.6) & 86.0 (3.5) & 82.3 (1.7) & 77.0 (4.0) & 85.9 (3.2)\\
BYOL           & 83.5 (0.1) & 78.2 (0.2) & 84.8 (0.5) & 84.3 (0.4) & 79.9 (0.8) & 90.5 (0.3) & 84.8 (1.0) & 82.7 (1.0) & 90.8 (0.3)\\
VICReg         & 82.1 (0.3) & 76.0 (0.4) & 80.7 (0.7) & 83.4 (0.8) & 77.8 (1.0) & 87.4 (0.5) & 82.1 (0.7) & 78.7 (1.9) & 88.0 (0.4)\\\hline
HiDisc-Patch   & 80.8 (0.0) & 73.5 (0.1) & 81.9 (0.0) & 82.3 (0.2) & 76.4 (0.2) & 88.3 (0.2) & 82.6 (0.3) & 77.9 (0.3) & 88.6 (0.2)\\
HiDisc-Slide   & 86.9 (0.2) & \tbf{83.2 (0.2)} & 87.4 (0.6) & \tbf{88.1 (0.5)} & 85.5 (0.3) & 91.9 (0.6) & \tbf{87.6 (0.5)} & \tbf{87.0 (1.4)} & \tbf{90.4 (1.4)}\\
HiDisc-Patient & \tbf{87.4 (0.1)} & \tbf{83.5 (0.2)} & \tbf{88.7 (0.2)} & \tbf{88.5 (0.2)} & \tbf{86.2 (0.2)} & \tbf{92.8 (0.2)} & \tbf{87.9 (0.5)} & \tbf{86.4 (0.6)} & \tbf{92.3 (1.1)}\\\hline
Supervised     & 88.9 (0.3) & 86.3 (0.3) & 90.8 (0.3) & 89.0 (0.5) & 88.8 (0.6) & 93.9 (0.3) & 88.5 (0.5) & 89.1 (0.5) & 93.6 (0.2)\\\hline
    \end{tabular}

    \caption{\textbf{Extended Main SRH Results.} Complete patch-, slide-, and patient-level metrics are shown. Slide-level metrics are aggregated using average pooling, similar to patient-level evaluation. Slide-level results are consistent with the patient-level metrics, showing HiDisc-Patient outperforms all other self-supervised learning baselines. We repeat experiments across three different random seeds, and standard deviations are reported in parentheses. MCA, mean class accuracy, AUPRC, area under the precision-recall curve.}
    \label{tab:main_srh_extended}
\end{table*}
\newcommand{\rclg}{\rowcolor{lg}}
\newcommand{\patchcolhead}[1]{\multirow{#1}{*}{\shortstack{\textbf{Patch}\\level\\metrics}}}
\newcommand{\slidecolhead}[1]{\multirow{-#1}{*}{\shortstack{\textbf{Slide}\\level\\metrics}}}
\newcommand{\patiecolhead}[1]{\multirow{#1}{*}{\shortstack{\textbf{Patient}\\level\\metrics}}}
\newcommand{\lastkhlinelg}[1]{\hhline{>{\arrayrulecolor{lg}}-*{#1}{>{\arrayrulecolor{black}}|-}|}}

\begin{table*}[ht!]
    \centering
    \setlength{\tabcolsep}{5pt}
    \begin{tabular}{c|c|cccccc}\hline
        & Method & Accuracy & MCA  & Sensitivity & Specificity & AUROC & AUPRC\\\hline
\patchcolhead{8}       & SimCLR         & 77.8 (0.0) & 77.5 (0.0) & 74.5 (0.2) & 80.5 (0.2) & 85.2 (0.1) & 79.6 (0.3)\\
                       & SimSiam        & 68.4 (0.3) & 68.0 (0.3) & 64.3 (0.6) & 71.7 (0.2) & 74.1 (0.3) & 65.4 (0.3)\\
                       & BYOL           & 80.0 (0.1) & 79.8 (0.1) & 77.7 (0.4) & 81.9 (0.3) & 87.5 (0.1) & 82.0 (0.3)\\
                       & VICReg         & 75.5 (0.1) & 75.2 (0.1) & 72.8 (0.4) & 77.6 (0.3) & 82.9 (0.1) & 75.5 (0.1)\\\cline{2-8}
                       & HiDisc-Patch   & 77.2 (0.1) & 76.7 (0.1) & 72.6 (0.1) & 80.9 (0.1) & 84.7 (0.1) & 78.8 (0.2)\\
                       & HiDisc-Slide   & 82.7 (0.2) & 82.5 (0.2) & 80.5 (0.2) & \tbf{84.4 (0.2)} & 89.3 (0.2) & 85.0 (0.2)\\
                       & HiDisc-Patient & \tbf{ 83.1 (0.1) } & \tbf{ 83.0 (0.1) } & \tbf{ 81.9 (0.3) } & \tbf{ 84.2 (0.2) } & \tbf{ 90.1 (0.1) } & \tbf{ 86.2 (0.2)}\\\cline{2-8}
                       & Supervised     & 85.1 (0.3) & 85.0 (0.3) & 83.7 (0.6) & 86.3 (0.1) & 91.7 (0.2) & 89.1 (0.2)\\\hline\hline
\rclg                  & SimCLR         & 83.0 (0.3) & 82.7 (0.3) & 80.1 (0.6) & \tbf{85.3 (0.3)} & 90.3 (0.2) & 86.5 (0.4)\\
\rclg                  & SimSiam        & 77.2 (0.7) & 76.6 (0.8) & 71.3 (1.2) & 81.9 (0.5) & 82.9 (0.4) & 73.4 (0.4)\\
\rclg                  & BYOL           & 84.1 (0.3) & 84.0 (0.3) & 83.5 (0.6) & \textbf{84.6 (0.5)} & 91.6 (0.2) & 88.2 (0.4)\\
\rclg                  & VICReg         & 80.8 (0.3) & 80.6 (0.3) & 78.3 (0.9) & 82.8 (0.6) & 88.7 (0.1) & 83.5 (0.3)\\\lastkhlinelg{7}
\rclg                  & HiDisc-Patch   & 82.7 (0.2) & 82.4 (0.2) & 79.4 (0.3) & \tbf{85.3 (0.3)} & 89.8 (0.2) & 85.2 (0.4)\\
\rclg                  & HiDisc-Slide   & \tbf{85.5 (0.4) } & \tbf{85.6 (0.4) } & \tbf{86.9 (0.9) } & 84.4 (0.3) & \tbf{93.9 (0.1) } & \tbf{91.7 (0.6)}\\
\rclg                  & HiDisc-Patient & \tbf{85.1 (0.2) } & \tbf{85.2 (0.2) } & \tbf{86.4 (0.3) } & 84.1 (0.3) & \tbf{93.8 (0.3) } & \tbf{91.7 (0.7)}\\\lastkhlinelg{7}
\rclg \slidecolhead{8} & Supervised     & 88.3 (1.3) & 88.3 (1.3) & 89.0 (1.7) & 87.7 (1.3) & 95.4 (0.1) & 94.5 (0.2)\\\hline\hline
\patiecolhead{8}       & SimCLR         & 80.7 (0.6) & 80.7 (0.6) & 80.2 (1.0) & \tbf{81.3 (0.7)} & 88.9 (0.3) & 86.1 (0.4)\\ 
                       & SimSiam        & 76.6 (0.6) & 76.5 (0.6) & 73.9 (1.1) & 79.0 (0.8) & 82.4 (0.3) & 73.3 (0.5)\\ 
                       & BYOL           & 83.1 (0.6) & 83.3 (0.6) & 86.6 (1.0) & 80.0 (1.1) & 89.8 (0.2) & 86.2 (0.6)\\ 
                       & VICReg         & 77.0 (0.5) & 77.0 (0.5) & 77.6 (1.0) & 76.3 (1.2) & 86.0 (0.3) & 79.9 (0.5)\\ \cline{2-8}
                       & HiDisc-Patch   & 81.0 (0.5) & 80.9 (0.5) & 79.6 (0.4) & \tbf{82.2 (0.9)} & 88.1 (0.2) & 84.1 (0.5)\\ 
                       & HiDisc-Slide   & \tbf{84.3 (0.3)} & \tbf{84.5 (0.3)} & \tbf{88.4 (0.8) } & 80.6 (0.7) & \tbf{92.3 (0.3) } & \tbf{89.4 (0.8)}\\ 
                       & HiDisc-Patient & \tbf{83.6 (0.3)} & 83.8 (0.3) & \tbf{87.6 (0.6) } & 80.0 (0.6) & \tbf{91.8 (0.2) } & \tbf{89.2 (0.8)}\\ \cline{2-8}
                       & Supervised     & 88.3 (0.4) & 88.4 (0.4) & 92.6 (0.8) & 84.3 (0.6) & 95.2 (0.2) & 94.2 (0.9)\\ \hline
    \end{tabular}
    \caption{\textbf{Extended Main TCGA Results.} For the binary molecular genetic classification task on TCGA dataset, we provide additional performance metrics, including sensitivity and specificity, as well as metrics at the whole-slide level. HiDisc maintains superior performance on all metrics across different levels compared to SSL baselines. We repeat experiments across three different random seeds, and randomly sampled 400 patches from each whole slide for nearest neighbor evaluation across three different random seeds. Standard deviations across nine evaluations are reported in parentheses. MCA, mean class accuracy, AUROC, area under the receiver operating characteristic curve, AUPRC, area under the precision-recall curve.}
    \label{tab:main_tcga_extended}
\end{table*}

\clearpage
\begin{table*}[ht!]
    \centering
    \setlength{\tabcolsep}{4pt}
    \begin{tabular}{c|ccc||>{\columncolor{lg}}c>{\columncolor{lg}}c>{\columncolor{lg}}c||ccc}\hline
    & \multicolumn{3}{c||}{\tbf{Patch} Level Metrics} & \multicolumn{3}{>{\columncolor{lg}}c||}{\tbf{Slide} Level Metrics} & \multicolumn{3}{c}{\tbf{Patient} Level Metrics}\\
    \hhline{>{\arrayrulecolor{white}}-*{3}{>{\arrayrulecolor{black}}|-}*{3}{>{\arrayrulecolor{black}}|-}*{3}{>{\arrayrulecolor{black}}|-}|}
        Method & Accuracy & MCA  & AUPRC& Accuracy & MCA  & AUPRC& Accuracy & MCA  & AUPRC\\\hline
SimCLR         & 31.5 (2.3) & 23.1 (1.9) & 25.0 (2.3) & 36.6 (4.1) & 28.5 (2.7) & 46.8 (4.3) & 40.2 (6.9) & 28.9 (4.5) & 48.4 (3.8)\\\hline
HiDisc-Patch   & 31.3 (0.6) & 22.2 (0.5) & 24.8 (1.1) & 43.0 (1.6) & 32.5 (1.1) & 50.9 (3.8) & 47.4 (2.1) & 33.1 (1.6) & 51.9 (2.3)\\
HiDisc-Slide   & 82.8 (0.2) & 77.4 (0.3) & 77.6 (0.3) & 85.5 (0.3) & \tbf{81.1 (0.5)} & 88.1 (0.2) & \tbf{84.2 (0.5)} & \tbf{82.3 (0.4)} & 88.6 (0.6)\\
HiDisc-Patient & \tbf{84.9 (0.2)} & \tbf{78.9 (0.1)} & \tbf{81.9 (0.3)} & \tbf{86.6 (0.2)} & \tbf{81.7 (0.7)} & \tbf{89.8 (0.1)} & \tbf{84.7 (0.5)} & \tbf{80.9 (1.4)} & \tbf{90.3 (0.2)}\\\hline
Supervised     & 90.0 (0.2) & 87.4 (0.3) & 85.4 (0.2) & 91.0 (0.3) & 90.7 (0.6) & 93.5 (0.8) & 90.0 (0.5) & 90.3 (0.4) & 93.2 (0.4)\\\hline
    \end{tabular}
    \caption{\textbf{Extended SRH Results with Weak Augmentations.} The weak augmentations here only use random vertical and horizontal flip. Without strong augmentation, the instance discrimination at patch level fails to provide a meaningful pretext task to learn meaningful representation. We only observe a 1-3 points drop as compared to strong augmentation on HiDisc-Slide and HiDisc-Patient, showing hierarchical discrimination reduces the reliance on the augmentation. We repeat experiments across three different random seeds, and standard deviations are reported in parentheses. MCA, mean class accuracy, AUPRC, area under the precision-recall curve.}
    \label{tab:weak_srh_extended}
\end{table*}

\begin{table*}[ht!]
    \centering
    \setlength{\tabcolsep}{5pt}
\begin{tabular}{c|c|cccccc}\hline
& Method & Accuracy & MCA  & Sensitivity & Specificity & AUROC & AUPRC\\\hline
\patchcolhead{5}        & SimCLR         & 57.1 (1.1) & 54.6 (1.0) & 31.2 (4.6) & 77.9 (4.7) & 58.4 (2.2) & 51.6 (1.9)\\\cline{2-8}
                        & HiDisc-Patch   & 59.0 (0.8) & 57.2 (0.8) & 40.0 (3.7) & 74.4 (3.2) & 61.5 (1.3) & 54.3 (1.0)\\
                        & HiDisc-Slide   & 79.6 (0.1) & 79.6 (0.1) & 79.0 (0.2) & 80.1 (0.1) & 86.3 (0.2) & 79.6 (0.4)\\
                        & HiDisc-Patient & \tbf{82.9 (0.2)} & \tbf{82.7 (0.2)} & \tbf{81.3 (0.2)} & \tbf{84.1 (0.2)} & \tbf{89.6 (0.2)} & \tbf{85.1 (0.4)}\\\cline{2-8}
                        & SupCon         & 85.4 (0.4) & 85.3 (0.4) & 83.9 (0.7) & 86.7 (0.2) & 92.0 (0.2) & 89.6 (0.5)\\\hline\hline
\rclg                   & SimCLR         & 58.7 (0.9) & 54.3 (1.0) & 12.8 (2.9) & 95.7 (2.0) & 70.2 (4.7) & 62.5 (2.7)\\\lastkhlinelg{7}
\rclg                   & HiDisc-Patch   & 63.7 (2.5) & 60.4 (2.9) & 29.7 (6.7) & \tbf{91.2 (2.1)} & 75.6 (3.4) & 67.9 (2.5)\\
\rclg                   & HiDisc-Slide   & 81.0 (0.3) & 81.0 (0.3) & 81.4 (0.7) & 80.6 (0.0) & 88.6 (0.2) & 83.4 (0.6)\\
\rclg                   & HiDisc-Patient & \tbf{85.0 (0.4)} & \tbf{85.2 (0.4)} & \tbf{86.9 (0.8)} & 83.5 (0.5) & \tbf{92.6 (0.1)} & \tbf{89.4 (0.7)}\\\lastkhlinelg{7}
\rclg \slidecolhead{5}  & SupCon         & 89.0 (0.7) & 89.1 (0.8) & 90.0 (1.2) & 88.1 (0.6) & 95.6 (0.2) & 94.8 (0.4)\\\hline\hline
\patiecolhead{5}        & SimCLR         & 58.1 (1.1) & 56.2 (1.1) & 14.9 (2.5) & 97.5 (1.2) & 72.8 (3.2) & 71.7 (2.0)\\\cline{2-8}
                        & HiDisc-Patch   & 61.2 (4.0) & 59.7 (4.2) & 25.3 (10.8) & \tbf{94.1 (3.4)}& 75.8 (2.5) & 71.8 (2.4)\\
                        & HiDisc-Slide   & 77.7 (0.4) & 77.7 (0.4) & 82.9 (0.8) & 72.8 (0.0) & 85.3 (0.3) & 79.5 (0.7)\\
                        & HiDisc-Patient & \tbf{82.3 (0.3)} & \tbf{82.5 (0.3)} & \tbf{86.6 (0.5)} & 78.4 (0.8) & \tbf{90.3 (0.3)} & \tbf{85.5 (1.3)}\\\cline{2-8}
                        & SupCon         & 88.4 (0.8) & 88.6 (0.8) & 92.7 (0.9) & 84.5 (0.7) & 95.2 (0.4) & 94.3 (0.7)\\\hline
\end{tabular}
    \caption{\textbf{Extended TCGA Results with Weak Augmentations.} The same weak augmentation experiment is conducted on TCGA dataset. Since this is a binary classification task, a random guess will have an accuracy of 50\%. We observe SimCLR and HiDisc-Patch have a low accuracy close to random guessing, suggesting they collapse without learning meaningful representation. We repeat experiments across three different random seeds, and randomly sampled 400 patches from each whole slide for nearest neighbor evaluation across three different random seeds. Standard deviations across nine evaluations are reported in parentheses. MCA, mean class accuracy, AUROC, area under the receiver operating characteristic curve, AUPRC, area under the precision-recall curve.}
    \label{tab:weak_tcga_extended}
\end{table*}

\clearpage

\begin{table*}[ht!]
    \centering
{
    \setlength{\tabcolsep}{4pt}
    \begin{tabular}{ccc|ccc||>{\columncolor{lg}}c>{\columncolor{lg}}c>{\columncolor{lg}}c||ccc}\hline
               &&& \multicolumn{3}{c||}{\tbf{Patch} Level Metrics} & \multicolumn{3}{>{\columncolor{lg}}c||}{\tbf{Slide} Level Metrics} & \multicolumn{3}{c}{\tbf{Patient} Level Metrics}\\
               \hhline{*{3}{>{\arrayrulecolor{white}}-}*{3}{>{\arrayrulecolor{black}}|-}*{3}{>{\arrayrulecolor{black}}|-}*{3}{>{\arrayrulecolor{black}}|-}|}
        $\lambda_\text{Patient}$ & $\lambda_\text{Slide}$ & $\lambda_\text{Patch}$ & Accuracy & MCA  & AUPRC& Accuracy & MCA  & AUPRC& Accuracy & MCA  & AUPRC\\\hline
        1 & 1 & 0 & \tbf{87.1} & \tbf{84.2} & 87.1 & \tbf{89.3} & \tbf{88.6} & \tbf{92.9} & \tbf{89.5} & \tbf{89.9} & \tbf{92.3}\\
        1 & 0 & 1 & 86.7 & 82.4 & 88.0 & 88.3 & 85.8 & 92.5 & 86.6 & 87.0 & 91.9\\
        0 & 1 & 1 & 86.7 & 81.6 & \tbf{88.1} & 88.0 & 84.4 & 92.4 & 87.6 & 83.0 & 92.2\\\hline
        1 & 1 & 5 & 86.8 & 81.8 & \tbf{88.7} & 87.1 & 82.4 & 92.2 & 86.6 & 82.1 & 91.9\\
        1 & 5 & 1 & \tbf{87.0} & 82.6 & 87.4 & \tbf{88.8} & 85.8 & \tbf{92.4} & \tbf{87.6} & 85.5 & 91.2\\
        5 & 1 & 1 & 86.8 & \tbf{83.7} & 86.6 & 88.4 & \tbf{87.6} & \tbf{92.4} & 86.1 & \tbf{86.4} & \tbf{92.4} \\\hline
    \end{tabular}
    }
    \caption{\textbf{Ablation study on $\mathbf{\lambda}$ weighting factor for the SRH dataset.} In these experiments, we changed one of the $\lambda$ coefficients to 0 or 5. HiDisc is relatively robust to changes in $\lambda_\text{Patient}$ and $\lambda_\text{Slide}$ values, as slide and patient level discrimination are complementary to each other. Interestingly, we can observe a slight performance drop when $\lambda_\text{Patch}$ is amplified, and removing patch discrimination slightly boosted performance. MCA, mean class accuracy, AUPRC, area under the precision-recall curve.}
    \label{tab:srh_tune_lambda}
\end{table*}

\begin{table*}[ht!]
    \centering
{
    \setlength{\tabcolsep}{4pt}
    \begin{tabular}{c|ccc|cccccc}\hline
       & $\lambda_\text{Patient}$ & $\lambda_\text{Slide}$ & $\lambda_\text{Patch}$ & Accuracy & MCA  & Sensitivity & Specificity & AUROC & AUPRC\\\hline
\patchcolhead{12} & 1 & 1 & 0 & \tbf{83.0 (0.1) } & \tbf{82.7 (0.1)}& 80.6 (0.1)  & \tbf{84.9 (0.1)} & 89.3 (0.1) & 84.7 (0.2)\\
                  & 1 & 0 & 1 & 82.8 (0.1) & \tbf{82.7 (0.1) } & \tbf{81.4 (0.0)} & 84.0 (0.1)  & \tbf{90.2 (0.1) } & \tbf{86.6 (0.1)}\\
                  & 0 & 1 & 1 & 82.4 (0.1) & 82.3 (0.1) & 80.8 (0.0) & 83.7 (0.2)  & \tbf{90.2 (0.1) } & \tbf{86.2 (0.3)}\\\cline{2-10}
                  & 1 & 1 & 2 & 82.8 (0.1) & 82.7 (0.1) & 81.3 (0.0) & 84.1 (0.2)  & \tbf{90.2 (0.1) } & \tbf{86.2 (0.2)}\\
                  & 1 & 2 & 1 & 83.1 (0.1) & \tbf{82.9 (0.1) }& 81.1 (0.1)  & \tbf{84.7 (0.1) } & \tbf{90.1 (0.1) } & \tbf{86.1 (0.2)}\\
                  & 2 & 1 & 1 & \tbf{83.3 (0.0)} & 83.2 (0.0) & \tbf{82.1 (0.0)} & 84.3 (0.1) & \tbf{90.1 (0.1) } & \tbf{86.1 (0.2)}\\\cline{2-10}
                  & 1 & 1 & 5 & 82.2 (0.1) & 82.0 (0.1) & 80.6 (0.1) & 83.4 (0.2) & 89.6 (0.1) & 85.5 (0.2)\\
                  & 1 & 5 & 1 & \tbf{83.0 (0.1) } & \tbf{82.8 (0.1) } & \tbf{81.2 (0.0) } & \tbf{84.4 (0.1) } & \tbf{90.0 (0.1) } & \tbf{86.2 (0.3)}\\
                  & 5 & 1 & 1 & \tbf{82.9 (0.0) } & \tbf{82.7 (0.0)} & 80.8 (0.1)& \tbf{84.6 (0.1)} & 89.5 (0.1) & 85.2 (0.2)\\\cline{2-10}
                  & 0 & 0 & 1 & 75.7 (0.1) & 75.1 (0.1) & 69.5 (0.2) & 80.8 (0.0) & 83.3 (0.1) & 76.8 (0.1)\\
                  & 0 & 1 & 0 & \tbf{83.0 (0.0) } & \tbf{82.8 (0.0) } & \tbf{80.6 (0.1) } & \tbf{84.9 (0.1) } & \tbf{89.7 (0.1) } & \tbf{85.7 (0.1)}\\
                  & 1 & 0 & 0 & 82.8 (0.1) & 82.6 (0.1) & \tbf{80.5 (0.2)} & 84.7 (0.0) & 89.1 (0.1) & 84.1 (0.1)\\\hline\hline
\rclg                   & 1 & 1 & 0 & 84.2 (0.2) & 84.3 (0.2) & 85.3 (0.4) & 83.2 (0.0) & \tbf{93.7 (0.1)} & \tbf{91.0 (0.9)}\\
\rclg                   & 1 & 0 & 1 & \tbf{85.5 (0.0)} & \tbf{85.7 (0.0)} & \tbf{87.0 (0.0)} & 84.3 (0.0) & \tbf{93.8 (0.1)} & \tbf{91.6 (0.4)}\\
\rclg                   & 0 & 1 & 1 & 84.6 (0.0) & 84.6 (0.0) & 84.4 (0.0) & \tbf{84.8 (0.0)} & \tbf{93.7 (0.2)} & \tbf{91.7 (0.4)}\\\lastkhlinelg{9}
\rclg                   & 1 & 1 & 2 & \tbf{85.1 (0.4)} & \tbf{85.2 (0.4)} & \tbf{86.1 (0.4)} & \tbf{84.3 (0.5)} & 93.6 (0.2) & \tbf{91.6 (0.5)}\\
\rclg                   & 1 & 2 & 1 & \tbf{85.1 (0.2)} & \tbf{85.2 (0.2)} & \tbf{86.1 (0.4)} & \tbf{84.3 (0.0)} & \tbf{93.7 (0.1)} & \tbf{91.4 (0.7)}\\
\rclg                   & 2 & 1 & 1 & 84.6 (0.0) & 84.8 (0.0) & \tbf{86.4 (0.0)} & 83.2 (0.0) & \tbf{93.8 (0.1)} & \tbf{91.4 (0.8)}\\\lastkhlinelg{9}
\rclg                   & 1 & 1 & 5 & 83.9 (0.2) & 83.9 (0.2) & 83.8 (0.0) & \tbf{83.9 (0.3)} & 93.1 (0.1) & 90.8 (0.4)\\
\rclg                   & 1 & 5 & 1 & \tbf{85.0 (0.2)} & \tbf{85.2 (0.2)} & \tbf{86.4 (0.0)} & \tbf{83.9 (0.3)} & \tbf{93.7 (0.1)} & \tbf{91.8 (0.4)}\\
\rclg                   & 5 & 1 & 1 & \tbf{84.9 (0.0)} & \tbf{85.1 (0.0)} & \tbf{86.4 (0.0)} & \tbf{83.8 (0.0)} & \tbf{93.8 (0.1)} & \tbf{91.6 (0.3)}\\\lastkhlinelg{9}
\rclg                   & 0 & 0 & 1 & 81.7 (0.8) & 81.2 (0.8) & 76.6 (1.1) & \tbf{85.9 (0.5)} & 89.0 (0.1) & 82.6 (0.2)\\
\rclg                   & 0 & 1 & 0 & \tbf{85.4 (0.2)} & \tbf{85.5 (0.2)} & \tbf{86.8 (0.4)} & 84.3 (0.0) & \tbf{93.8 (0.1)} & \tbf{91.6 (0.7)}\\
\rclg \slidecolhead{12} & 1 & 0 & 0 & \tbf{85.6 (0.2)} & \tbf{85.7 (0.2)} & \tbf{86.8 (0.4)} & 84.6 (0.3) & 93.6 (0.0) & \tbf{90.1 (0.6)}\\\hline\hline
\patiecolhead{12}   & 1 & 1 & 0 & 81.4 (0.3) & 81.6 (0.3) & 84.9 (0.7) & 78.3 (0.0) & 91.6 (0.0) & \tbf{87.9 (0.9)}\\
                    & 1 & 0 & 1 & \tbf{84.1 (0.0)} & \tbf{84.3 (0.0)} & \tbf{88.1 (0.0)} & 80.4 (0.0) & \tbf{91.9 (0.2)} & \tbf{89.3 (0.7)}\\
                    & 0 & 1 & 1 & 83.7 (0.3) & 83.8 (0.3) & 86.1 (0.7) & \tbf{81.5 (0.0)} & \tbf{91.5 (0.3)} & \tbf{89.3 (0.9)}\\\cline{2-10}
                    & 1 & 1 & 2 & \tbf{83.3 (0.9)} & \tbf{83.5 (0.9)} & 86.5 (0.7) & \tbf{80.4 (1.1)} & 91.4 (0.3) & \tbf{88.9 (0.8)}\\
                    & 1 & 2 & 1 & \tbf{83.9 (0.3)} & \tbf{84.1 (0.3)} & \tbf{87.7 (0.7)} & \tbf{80.4 (0.0)} & 91.4 (0.1) & \tbf{88.2 (0.5)}\\
                    & 2 & 1 & 1 & 83.0 (0.0) & 83.2 (0.0) & \tbf{88.1 (0.0)} & 78.3 (0.0) & \tbf{92.0 (0.1)} & \tbf{88.6 (0.8)}\\\cline{2-10}
                    & 1 & 1 & 5 & 82.0 (0.3) & 82.1 (0.3) & 84.5 (0.0) & \tbf{79.7 (0.6)} & 90.9 (0.2) & \tbf{88.4 (0.5)}\\
                    & 1 & 5 & 1 & 83.1 (0.3) & 83.3 (0.3) & 86.9 (0.0) & \tbf{79.7 (0.6)} & \tbf{92.0 (0.1)} & \tbf{89.2 (0.5)}\\
                    & 5 & 1 & 1 & \tbf{83.5 (0.0)} & \tbf{83.7 (0.0)} & \tbf{88.1 (0.0)} & \tbf{79.3 (0.0)} & 91.6 (0.2) & \tbf{88.5 (0.5)}\\\cline{2-10}
                    & 0 & 0 & 1 & 80.7 (1.7) & 80.5 (1.7) & 77.4 (2.4) & \tbf{83.7 (1.1)} & 87.4 (0.2) & 81.2 (0.4)\\
                    & 0 & 1 & 0 & \tbf{83.5 (0.0)} & \tbf{83.7 (0.0)} & \tbf{86.9 (0.0)} & 80.4 (0.0) & \tbf{92.0 (0.1)} & \tbf{88.9 (0.7)}\\
                    & 1 & 0 & 0 & \tbf{83.7 (0.3)} & \tbf{83.8 (0.3)} & \tbf{86.5 (0.7)} & 81.2 (0.6) & 91.2 (0.0) & 85.9 (0.3)\\\hline
    \end{tabular}
    }
    \caption{\textbf{Ablation study on $\mathbf{\lambda}$ weighting factor for the TCGA dataset.} In these experiments, we changed one of the $\lambda$ coefficients to 0, 2or 5, as well as changing two of the $\lambda$ coefficients to 0. We randomly sample 400 patches from each whole slide for nearest neighbor evaluation across three different random seeds, and standard deviations are reported in parentheses. We can observe that HiDisc is relatively robust to changes in $\lambda_\text{Patient}$ and $\lambda_\text{Slide}$ values. As expected, when $\lambda_\text{Patient}=\lambda_\text{Slide}=0$, we observe a reduction in model performance because only patch discrimination is used to supervise model training. MCA, mean class accuracy, AUROC, area under the receiver operating characteristic curve, AUPRC, area under the precision-recall curve.}
    \label{tab:tcga_tune_lambda}
\end{table*}

\clearpage

\begin{table*}[ht!]
    \centering
{
    \setlength{\tabcolsep}{4pt}
    \begin{tabular}{c|ccc||>{\columncolor{lg}}c>{\columncolor{lg}}c>{\columncolor{lg}}c||ccc}\hline
               & \multicolumn{3}{c||}{\tbf{Patch} Level Metrics} & \multicolumn{3}{>{\columncolor{lg}}c||}{\tbf{Slide} Level Metrics} & \multicolumn{3}{c}{\tbf{Patient} Level Metrics}\\\hhline{>{\arrayrulecolor{white}}-*{3}{>{\arrayrulecolor{black}}|-}*{3}{>{\arrayrulecolor{black}}|-}*{3}{>{\arrayrulecolor{black}}|-}|}
        LR & Accuracy & MCA  & AUPRC & Accuracy & MCA  & AUPRC & Accuracy & MCA  & AUPRC\\\hline
        1    & 63.9 & 52.2 & 57.2 & 68.5 & 55.8 & 67.4 & 67.0 & 53.8 & 72.8\\
        1E-1 & 84.2 & 78.3 & 84.0 & 85.4 & 80.6 & 88.6 & 85.2 & \tbf{83.1} & 89.3\\
        1E-2 & 85.0 & \tbf{79.7} & 82.5 & \tbf{86.9} & \tbf{82.3} & 89.5 & \tbf{86.6} & 82.1 & 88.9\\
        1E-3 & 85.2 & 79.0 & 83.8 & 86.4 & 80.4 & \tbf{90.3} & 84.2 & 79.5 & \tbf{90.7}\\
        1E-4 & \tbf{85.5} & 78.7 & \tbf{84.5} & 86.2 & 80.6 & 89.6 & 85.6 & 80.6 & 89.8\\
        1E-5 & 85.2 & 78.2 & 84.0 & 85.4 & 77.4 & 89.4 & 84.2 & 77.2 & 89.9\\
        1E-6 & 78.8 & 69.9 & 76.4 & 80.5 & 72.1 & 82.9 & 81.8 & 74.5 & 85.9\\
        1E-7 & 68.5 & 58.5 & 64.8 & 75.0 & 65.9 & 73.1 & 74.2 & 65.0 & 74.8\\\hline
    \end{tabular}
    }
    \caption{\textbf{SRH Learn rate ablations.} We can observe that HiDisc training is robust to learning rate variations, achieving good performance from $10^{-1}$ to $10^{-5}$ in SRH. These experiments are performed with weak augmentations. MCA, mean class accuracy, AUPRC, area under the precision-recall curve.}
    \label{tab:srh_tune_lr}
\end{table*}

\begin{table*}[ht!]
    \centering
    \setlength{\tabcolsep}{5pt}
\begin{tabular}{c|c|cccccc}\hline
& LR & Accuracy & MCA  & Sensitivity & Specificity & AUROC & AUPRC\\\hline
\patchcolhead{3}        & 0.01    & \tbf{83.1 (0.1)} & \tbf{82.9 (0.1)} & 81.3 (0.1) & \tbf{84.6 (0.1)} & \tbf{89.9 (0.1)} & 85.5 (0.1)\\
                        & 0.001   & \tbf{83.1 (0.1)} & \tbf{82.9 (0.1)} & \tbf{81.9 (0.1)} & 84.0 (0.1) & \tbf{90.0 (0.1)} & \tbf{86.1 (0.1)}\\
                        & 0.0001  & 82.0 (0.2) & 81.9 (0.1) & 81.1 (0.1) & 82.7 (0.2) & 89.5 (0.1) & 85.4 (0.2)\\\hline
\rclg                   & 0.01    & \tbf{85.8 (0.0)} & \tbf{85.9 (0.0)} & \tbf{87.0 (0.0)} & \tbf{84.8 (0.0)} & \tbf{93.6 (0.1)} & \tbf{91.4 (0.4)}\\
\rclg                   & 0.001   & 85.0 (0.2) & 85.2 (0.2) & 86.4 (0.0) & 83.9 (0.3) & \tbf{93.5 (0.1)} & \tbf{91.2 (0.8)}\\
\rclg \slidecolhead{3}  & 0.0001  & \tbf{85.2 (0.8)} & \tbf{85.3 (0.8)} & 85.9 (1.4) & \tbf{84.6 (0.3)} & 93.2 (0.2) & \tbf{90.6 (0.4)}\\\hline
\patiecolhead{3}        & 0.01    & \tbf{84.1 (0.0)} & 84.2 (0.0) & 86.9 (0.0) & 81.5 (0.0) & \tbf{91.5 (0.1)} & \tbf{88.4 (0.8)}\\
                        & 0.001   & 83.7 (0.3) & 83.9 (0.3) & \tbf{88.1 (0.0)} & 79.7 (0.6) & \tbf{91.6 (0.2)} & \tbf{88.6 (0.5)}\\
                        & 0.0001  & \tbf{84.8 (0.7)} & \tbf{85.0 (0.7)} & 87.7 (0.7) & \tbf{82.2 (0.6)} & 90.8 (0.2) & 87.9 (0.5)\\\hline
\end{tabular}
    \caption{\textbf{TCGA learn rate ablations}. We can observe that HiDisc training is robust to learning rate variations, achieving good performance from $10^{-4}$ to $10^{-2}$ in TCGA. These TCGA experiments are performed with strong augmentations, and 400 patches are sampled randomly from each whole slide for nearest neighbor evaluation across three different random seeds, and standard deviations are reported in parentheses. MCA, mean class accuracy, AUROC, area under the receiver operating characteristic curve, AUPRC, area under the precision-recall curve.}
    \label{tab:tcga_tune_lr}
\end{table*}

\begin{table*}[ht!]
    \centering
{
    \setlength{\tabcolsep}{4pt}
    \begin{tabular}{c|ccc||>{\columncolor{lg}}c>{\columncolor{lg}}c>{\columncolor{lg}}c||ccc}\hline
               & \multicolumn{3}{c||}{\tbf{Patch} Level Metrics} & \multicolumn{3}{>{\columncolor{lg}}c||}{\tbf{Slide} Level Metrics} & \multicolumn{3}{c}{\tbf{Patient} Level Metrics}\\\hhline{>{\arrayrulecolor{white}}-*{3}{>{\arrayrulecolor{black}}|-}*{3}{>{\arrayrulecolor{black}}|-}*{3}{>{\arrayrulecolor{black}}|-}|}
        Effective batch size & Accuracy & MCA  & AUPRC & Accuracy & MCA  & AUPRC & Accuracy & MCA  & AUPRC\\\hline
512  & \tbf{85.8} & \tbf{81.3} & \tbf{87.6} & \tbf{88.1} & \tbf{85.1} & \tbf{92.6} & \tbf{88.0} & \tbf{85.7} & \tbf{92.7}\\
1024 & \tbf{85.8} & 80.4 & 87.4 & 87.1 & 83.4 & 92.2 & \tbf{88.0} & 85.4 & 92.3\\\hline
    \end{tabular}
    }
    \caption{\textbf{Batch size ablations.} We perform ablation studies to investigate the effect of batch size on HiDisc training. Due to the computation resources limit, we are only able to ablate batch size on 512 and 1024 for SRH dataset. We can observe that HiDisc training does not benefit from a larger batch size. Experiments in the table are performed without sync batch norm. MCA, mean class accuracy, AUPRC, area under the precision-recall curve.}
    \label{tab:srh_tune_bs}
\end{table*}

\end{document}